\documentclass[10pt,twocolumn,letterpaper]{article}

\usepackage{wacv}              % To produce the CAMERA-READY version
\usepackage{graphicx}
\usepackage{amsmath}
\usepackage{amssymb}
\usepackage{booktabs}
\usepackage{subcaption}
\usepackage{txfonts}
\usepackage[pagebackref,breaklinks,colorlinks]{hyperref}

\usepackage[accsupp]{axessibility}  % Improves PDF readability for those with disabilities.
\usepackage[capitalize]{cleveref}
\crefname{section}{Sec.}{Secs.}
\Crefname{section}{Section}{Sections}
\Crefname{table}{Table}{Tables}
\crefname{table}{Tab.}{Tabs.}

\def\wacvPaperID{1257} % *** Enter the WACV Paper ID here
\def\confName{WACV}
\def\confYear{2024}

\begin{document}

\title{TSP-Transformer: Task-Specific Prompts Boosted Transformer \\for Holistic Scene Understanding}

\author{Shuo Wang$^{1}$\quad
Jing Li$^{2}$ \quad
Zibo Zhao$^{1}$ \quad
Dongze Lian$^{3}$ \\
Binbin Huang$^{1}$ \quad
Xiaomei Wang$^{4}$ \quad
Zhengxin Li$^{1}$ \quad
Shenghua Gao$^{1,5,6}$\footnote[2]{} \quad
\\
$^{1}$ShanghaiTech University \quad
$^{2}$Xiaohongshu Inc.\quad
$^{3}$National University of Singapore \\
$^{4}$Fudan University\quad
$^{5}$Shanghai Engineering Research Center of Intelligent Vision and Imaging\\
$^{6}$Shanghai Engineering Research Center of Energy Efficient and Custom AI IC\\
{\tt\small   \{wansghuo2022, zhaozb, liandz, huangbb, lizhx, gaoshh\}@shanghaitech.edu.cn}\\ {\tt\small lijing1@alumni.shanghaitech.edu.cn, 17110240025@fudan.edu.cn}\\
}

\maketitle
%%%%%%%%% ABSTRACT
\begin{abstract}
Holistic scene understanding includes semantic segmentation, surface normal estimation, object boundary detection, depth estimation, etc. The key aspect of this problem is to learn representation effectively, as each subtask builds upon not only correlated but also distinct attributes. Inspired by visual-prompt tuning, we propose a \textbf{Task-Specific Prompts Transformer}, dubbed TSP-Transformer, for holistic scene understanding. It features a vanilla transformer in the early stage and tasks-specific prompts transformer encoder in the lateral stage, where tasks-specific prompts are augmented. By doing so, the transformer layer learns the generic information from the shared parts and is endowed with task-specific capacity. First, the tasks-specific prompts serve as induced priors for each task effectively. Moreover, the task-specific prompts can be seen as switches to favor task-specific representation learning for different tasks.
Extensive experiments on NYUD-v2 and PASCAL-Context show that our method achieves state-of-the-art performance, validating the effectiveness of our method for holistic scene understanding. We also provide our code in the following link \footnote{\url{https://github.com/tb2-sy/TSP-Transformer}}.

\end{abstract}

\section{Introduction}

\begin{figure}[h]
\includegraphics[width=8cm]{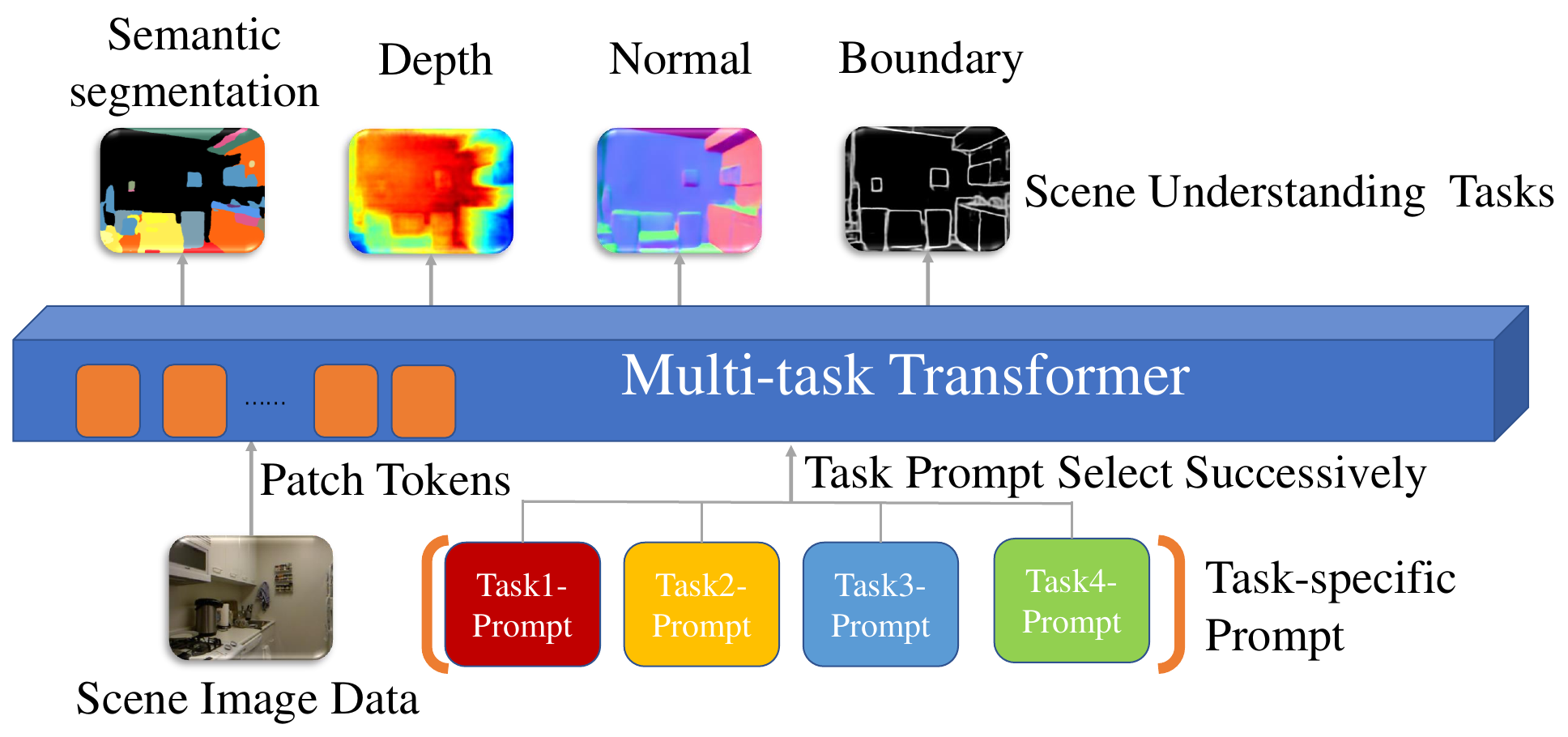}
        %\captionsetup{belowskip=-10mm}
	\caption{Overview of our proposed Task-Specific Prompts Transformer Network framework. We develope task-specific prompts for each task, which successively interact with the encoder to generate features relevant to the respective tasks. Task-specific prompts boost Multi-task Transformer with task-specific features and output the final multi-task predictions.}
  \vspace{-7mm}
	\label{fig:framework}
\end{figure}
%%\vspace{-5mm}

Understanding a scene \cite{uncertainty,1,li2022learning} both semantically and geometrically is promising due to its lots of applications, such as autonomous driving, virtual reality, robotics, etc. The holistic understanding of the scene includes a wide variety of tasks, e.g., semantic segmentation, depth estimation, object boundary detection, surface normal estimation, etc. It is desirable to train a network that can deal with a series of tasks jointly. By leveraging the task association and task specialty, impressive results on scene understanding are achieved by employing Convolution Neural Networks (CNNs) \cite{PAD-Net,mti-Net,atrc,ubernet}. Recently, transformer models \cite{invpt,bai2021transformers,hrformer,pyramid} have been introduced to model the multiple scene understanding tasks.

Previous works have designed different modules to learn task-specific and task-unified knowledge and interact with the information of these various tasks. For example, lots of works have designed special modules in the encoder \cite{end2end,nddr,crossstitch} to learn task-specific representations and incorporate cross-task information interactions by means of manually designed structures.
%which enable the seamless integration of relevant features and knowledge across different tasks.
% embed cross-task information interactions through hand-crafted structures 
While the decoder \cite{invpt,universal,mti-Net} serves to decouple the characteristics of different tasks and facilitate cross-task interaction, some parameters are shared across all tasks. However, these methods are not fine enough for decoupling task-specific features and require carefully designed modules for multi-task dense learning. 

Recently, building a unified network to perform multiple tasks has received more attention due to its potential for general intelligence. A large number of transformer-based frameworks \cite{bert,gpt3,jiang2020can} have been constructed in the field of natural language processing to handle many different but related tasks.
% A typical application is ChatGPT, which can perform your task based on the input instructions. 
In computer vision, some recent approaches \cite{unified_interface} design unified models \cite{opportunities} to obtain the output results based on the task description by implicitly learning the relevance of tasks, which show the feasibility of a unified model given instructions to perform multiple vision tasks. Inspired by these unified models \cite{unified_interface,unifytoken}, we propose to utilize multi-task visual prompts to improve the multi-task scene understanding. Each 
% group of
prompt is dedicated to a specific task, which can be regarded as task-specific. These multi-task visual prompts are inserted into the input of the transformer layer along with the image tokens to perform self-attention such that the transformer layer learns the task-specific information. Such a method decouples characteristics of different tasks in the layer level, which is finer and more flexible. 
% To the best of our knowledge, we are the pioneers of applying task-specific prompts to the multi-task scene understanding area. 

% from the shared parts and is endowed with task-specific capacity. 

Different from the previous unified models \cite{unified_interface}, where the explicit task description is inserted into the transformer to specify the task name, \emph{e.g.}, `detect' or `segment' , our task-specific prompts are learnable and updated independently for each task during training.
% , and other parameters of the transformer layer are shared across multiple tasks. 
Thanks to these learned task-specific prompts, the proposed model demonstrates the ability to produce task-specific output, which provides tailored results for specific requirements and is better adapted to the data distribution, and largely improves the performance. Furthermore, we also introduce an efficient task-specific feature fusion module to aggregate the features from the encoder of different tasks and pass them to the decoder.

We conduct extensive experiments on NYUD-v2 and PASCAL-Context for performance evaluation. %Our experiments reveal that only task-specific features should be inserted into the deeper layers of the encoder because different tasks tend to have similar low-level features, and as we gradually introduce more and more task-specific prompts in deeper encoder layers, features of different tasks become less and less correlated, which helps task-specific feature learning.%
Our experiments reveal that when task-specific features are inserted into the deeper layers of the encoder, then different tasks tend to have similar low-level features, and as we gradually introduce more and more task-specific prompts in deeper encoder layers, features of various tasks become less and less correlated, which helps task-specific features learning. Further, although tasks are closely correlated, our experiments show that introducing unified prompts is adverse to performance. The experiment results demonstrate that our proposed method is superior to other baseline methods, which validates the effectiveness of our proposed approach. In summary, our main contributions are as follows:
\begin{itemize}
    \item We are introducing a novel approach, Task-Specific Prompts Transformer, to include task-specific prompts and various task features in multi-task training of scene understanding.
    \item We explore a new task-specific feature decoupling method; task-specific information can be finely decoupled with task-specific prompts. And we also design an efficient task-specific encoder feature fusion block to aggregate the encoder features of different tasks.
    \item We conduct extensive experiments on two benchmarks: NYUD-v2 and PASCAL-Context. The experiment results demonstrate that our proposed method is superior to other baseline methods. 

\end{itemize}

% 1. 

% 2. We explore a new task-specific feature decoupling method; task-specific information can be finely decoupled with task-specific prompts. And we also design an efficient task-specific encoder feature fusion block to aggregate the encoder features of different tasks.

% 3. We conduct extensive experiments on two benchmarks: NYUD-v2 and PASCAL-Context. The experiment results demonstrate that our proposed method is superior to other baseline methods. 

% %\vspace{-0.4mm}
%%%\vspace{-0.4cm}
\section{Related Work}
\label{sec:related_work}
\noindent\textbf{Multi-task learning for dense scene understanding.}
Multi-task learning (MTL) is a promising area \cite{1,uncertainty,lian2018multiview} of research to improve generalization performance by using domain knowledge contained in supervised signals of associated tasks. 
% Multi-task learning is more effective when there is an auxiliary role \cite{PAD-Net} between related tasks and can achieve better results. 
Many previous works \cite{PAD-Net,mti-Net,atrc,invpt} have explored lots of possibilities in this field. In particular, PAD-Net \cite{PAD-Net} proposes a new multi-task-oriented predictive distillation network structure that supplies abundant multi-modal data for learning the target task. MTI-Net \cite{mti-Net} explicitly considers task interactions at multiple scales by designing the network structure and utilizing multi-scale information. Multi-scale processing preserves the original image’s features across various scales, such as the patch level. 
Neural architecture search (NAS) \cite{darts,snas,lian2020iclr} techniques have also been employed in the field of multi-task learning, such as ATRC \cite{atrc}, with the help of knowledge distillation techniques to obtain richer information in the case of multi-task information sharing. The above methods are almost implemented based on CNNs. Owing to the exceptional performance of transformers in the visual domain, many researches \cite{invpt} are being explored using Transformers \cite{attention}. 
% Vision Transformer \cite{vit} approach involves dividing the image into patch level and converting them into tokens. This operation enables the direct application of Transformers to the images domain, originally designed for the Nature Language Process field. 
Based on the structure of the Vision Transformer, InvPT \cite{invpt} improves the decoder to expand the receptive field and realizes information interaction at higher resolutions. In this paper, our work explores the role of the Visual Prompt Transformer in multi-task learning and proposes a new method to use Visual Prompt Learning \cite{vpt} in multi-task learning.

\noindent\textbf{Visual prompt learning.}
Prompt learning first emerged in the field of natural language processing \cite{shin2020autoprompt,petroni2019language,lester2021power,li2021prefix,liu2021p} for parameter-efficient tuning, which enables the model to better understand tasks and improve performance. 
Some large models in the field of natural language processing, such as GPT-3 \cite{gpt3}, have demonstrated the ability to transfer under small sample conditions under the blessing of the prompt. 
% Recent work has extended prompt to the field of computer vision\cite{vpt,zhou2022conditional,ju2022prompting,radford2021learning},
Recent work has extended the prompt learning or parameter-efficient learning to the field of computer vision \cite{vpt,wu2022generative,sandler2022fine,zhou2022conditional,xu2022match,Lian_2022_SSF}, and the modality of the prompt is not limited to text. VPT \cite{vpt} inserts some learnable parameters into the Vision Transformer Encoder Layer, and these learnable token parameters interact with the input image token, thereby affecting downstream tasks. However, the prompts method \cite{vpt} mentioned above in the visual field are limited to fine-tune classification tasks and have not been tried in other settings, such as multi-task learning. 

Recently, we noticed that there are some other concurrent works with similar ideas yet distinct methods, such as \cite{liang2023visual,ye2023taskprompter}. Although both our model and TaskPrompter \cite{ye2023taskprompter} enhance model performance for multi-task scene understanding with the proposed prompts to guide the learning process, our implementation methods and TaskPrompter have their own advantages and disadvantages. TaskPrompter uses two prompting paradigms in spatial and channel, called Spatial Task Prompting and Channel Task Prompting. They design a set of spatial-channel task prompts and learn their spatial and channel interactions with the shared image tokens in each transformer layer with attention mechanism. However, our method uses the standard visual prompt paradigm \cite{vpt}. The self-attention mechanism generates task-specific features by sequentially concating the learnable prompts of different tasks and image patch tokens. Task-specific features and task fusion features are used as the input of the decoder stage.

%%%%\vspace{-10mm}
\begin{figure*}[t]
	\includegraphics%[width=20cm]
	[scale=0.6]{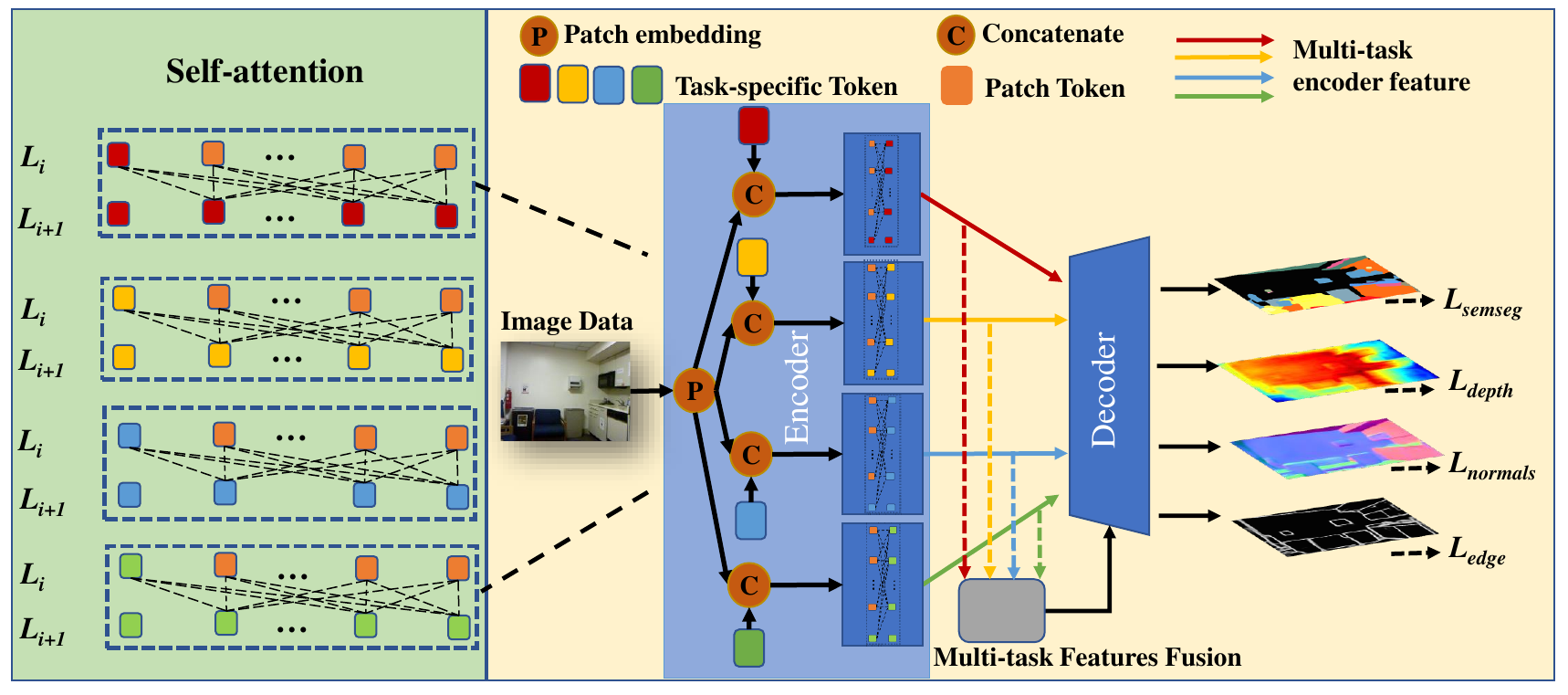}
         \captionsetup{belowskip=-12pt}
	\caption{Pipeline of our proposed Task-Specific Prompts Transformer. It consists of three parts: Task-specific prompts, Encoder, and Decoder. Given an image and a specific task, the task prompts are adopted to concatenate with the image feature tokens and perform self-attention in the Encoder. Successively perform the above steps to generate multi-task features from the Encoder. The features output from the Encoder are fused and fed into the Decoder to perform multi-task prediction.}
	%\label{fig:galaxy}%what use?
	\label{fig1}
\end{figure*}

\section{Method}

\subsection{Overview}

In this section, we introduce our Task-Specific Prompts Transformer Network. The overall framework can be shown in \cref{fig1}. Our method is implemented on \cite{invpt}, consisting of three core parts, Task-Specific Prompts Transformer Encoder in \cref{sec:3_2}, Efficient Multi-task Feature Fusion block in \cref{sec:3_3} and Multi-head Transformer Decoder in \cref{sec:3_4}. The details of these parts are introduced as follows. Task-Specific Prompts Transformer Encoder consists of vanilla transformer encoder layers \cite{attention} in the early stage and our proposed task-specific prompts transformer encoder layers in the late stage. A few shared vanilla transformer encoder layers in the early stage can reduce the number of parameters and computations and learn general knowledge. Task-specific prompts transformer encoder layers decouple different task features and learn task-specific information. Task-specific prompts are introduced to the task-specific prompts transformer encoder layers interacts with the image feature to learn corresponding task dependencies. 
% Prompt transformer encoder layers can
% achieve the decoupling of different task features and learn task-specific information.
Efficient Multi-task Feature Fusion block aggregates fundamental encoder features from different tasks to pass to the decoder stage, which is vital for some low-level tasks. Multi-head Transformer decoder learns to produce refined scene representations within global spatial and task information, which are further used to produce the final predictions with task-specific output head.

% The multi-task prompt transformer encoder consists of regular transformer encoder layers and our proposed multi-task prompt transformer layers. Our task-specific prompts are concatenate to the transformer encoder layer since a certain layer and interacts with the image feature to learn corresponding task dependencies. Efficient Encoder Multi-task Features Fusion block aggregates fundamental encoder features from different tasks to pass to decoder stage which is vital for some low level tasks.
% InvPT Transformer decoder learn to produce refined scene representations within global spatial and task contexts, which are further used to produce the final predictions with task-specific output head.
\subsection{Task-Specific Prompts Transformer Encoder}
\label{sec:3_2}
As shown in \cref{fig1}, our Task-Specific Prompts Transformer Encoder mainly consists of three parts, patch embedding, shared vanilla transformer encoder layers in the early stage, and task-specific prompts transformer encoder layers in the late stage. A few shared vanilla transformer layers in the early stage of the encoder, and task-specific prompts are introduced from particular layers in our default model. A few shared encoder layers in the early stage can reduce the number of parameters and computations and learn general knowledge. Task-specific prompts encoder layers in the late stage decouple different task features and learn task-specific information, and this decoupling is finely performed at the layer level.%The Transformer encoder acts as a backbone and is shared by all tasks which learning task-unify representation. While Task-specific prompts learn task-relevant features by interacting with the part of task-unify representation. %and  share information from different tasks with cross prompt attention module.

\noindent\textbf{Patch embedding.}
For a vanilla Vision Transformer (ViT) architecture with N layers, an input image $I \in \mathbb{R}^{H \times W \times 3}$ is divided into $m$ fixed-sized patches $I_j$. Each image patch $I_j$ is then first embedded into a D-dimensional embedding token with positional encoding:
%\vspace{-0.1cm}
\begin{equation}
\mathbf{e}_0^j=\operatorname{Embed}\left(I_j\right) \quad \mathbf{e}_0^j \in \mathbb{R}^D, j=1,2, \ldots m
\end{equation}
The embeddings of the $j$-th image patch after embedded are denoted as $\mathbf{e}_0^j$ ($\mathbf{e}_0^j \in \mathbb{R}^D, j=1,2, \ldots m$). We use $\mathbf{E}_i=\operatorname{Concat}\left(\mathrm{e}_i^1, \ldots, \mathrm{e}_i^j\right)$ as the collection of image patch token embeddings.

%, where $\operatorname{Concat(.)}$ denotes the concatenation operation. 

\noindent\textbf{Shared vanilla transformer encoder layer.}
%task-specific prompt replace intermediate supervision %
After image patch embedding, $\mathbf{E}_i$ is then input to the ($i+1$)-th Transformer encoder layer ${L}_{i+1}$:
%\vspace{-0.15cm}
\begin{equation}
\left[\mathbf{x}_{i+1}, \mathbf{E}_{i+1}\right]=L_{i+1}\left(\left[\mathbf{x}_{i}, \mathbf{E}_{i}\right]\right) \quad i=0,1,2, \ldots, M-1
\end{equation}
where $\mathbf{x}_i\in \mathbb{R}^{D}$ denote learnable [CLS]’s embedding at ${L}_{i}$ transformer encoder layer output space.

The shared vanilla transformer encoder layer $L$ consists of two main blocks of multi-head self-attention (MSA) and multi-layer perceptron (MLP). Given an input  $ \mathbf{E}_i\in \mathbb{R}^{m \times D}$, where $D$ is the embedding dimension, MSA first maps $ \mathbf{E}_i$ to queries $ \mathbf{Q}\in \mathbb{R}^{m \times d}$, keys $ \mathbf{K}\in \mathbb{R}^{m \times d}$ and values $ \mathbf{V}\in \mathbb{R}^{m \times d}$ using three projection matrices, $ \mathbf{W}_q\in \mathbb{R}^{D \times d}$, $ \mathbf{W}_k\in \mathbb{R}^{D \times d}$ and $ \mathbf{W}_v\in \mathbb{R}^{D \times d}$, where $d$ denote the hidden layer dimension. And MSA computes the weighted sums over the values based on the self-attention between the queries and keys as follows:
%\vspace{-0.4cm}
\begin{equation}
\operatorname{Attention}(Q, K, V)=\operatorname{softmax}\left(\frac{Q K^T}{\sqrt{d}}\right) V
\end{equation}
where $\frac{\mathit{1}}{\mathit{\sqrt{d}}}$ is a scaling factor.

\noindent\textbf{Task-specific prompts transformer encoder layer.}
We concatenate $n$ task-specific prompt tokens $p \in \mathbb{R}^{D}$ for each task which are initialized with a fixed value or a random distribution to the Task-specific prompts transformer encoder Layer for each task to learn task-specific features. Task-specific prompts forward separately and interact with the patch token embedding features through MSA to learn dependencies for different tasks. Task-specific prompts are introduced at the input space of particular transformer layers, and distinct layers introduced may yield varying performance. For ($M+1$)-th Layer ${L}_{M+1}$, we denote the collection of task $t$ input learnable task-specific prompts as $\mathbf{P}_0^t=\left\{\mathbf{p}_0^k \in \mathbb{R}^D \mid k \in \mathbb{N}, 1 \leq k \leq n\right\}$, denoted $t=1,2, \ldots, T$, where $T$ is the number of tasks. The task-specific prompts transform encoder layer is formulated as:

\begin{equation}
\begin{aligned}
% \left[\mathbf{x}_1,\mathbf{E}_1\right]&=L_1\left(\left[\mathbf{x}_{0}, \mathbf{E}_{0}\right]\right) \\
% \left[\mathbf{x}_2,\mathbf{E}_2\right]&=L_2\left(\left[\mathbf{x}_{1}, \mathbf{E}_{1}\right]\right) \\
% &\ldots \\
\left[\mathbf{x}^{t}_{M+1},\mathbf{E}^{t}_{M+1},\_ \right]&=\quad 
 L_{M+1}\left(\left[\mathbf{x}_{M}, \mathbf{E}_{M},\mathbf{P}_{0}^t\right]\right) \\
&\ldots \\
\left[\mathbf{x}^{t}_N,\mathbf{E}^{t}_N,\_ \right]\quad
&=L_N\left(\left[\mathbf{x}^{t}_{N-1}, \mathbf{E}^{t}_{N-1},\mathbf{P}_{N-M-1}^t\right]\right) \\
%&\quad \quad \quad i=1,2, \ldots, N
\end{aligned}
\end{equation}

where '$\_$' means tokens are discarded and replaced by next-layer task-specific prompt tokens. If we choose to introduce task-specific prompt tokens $\mathbf{P}_{0}^t$ in the $i$-th encoder layer, then $\mathbf{P}_{0}^t$ concatenate of image patch token embedding ${E}_{i-1}$ and learnable [CLS]’s embedding $\mathbf{x}_{i-1}$. Then, patch token embeddings after task-specific prompt interactions turn into task-specific features from encoder. In this way, task-specific prompts encoder layers achieve the decoupling of different task features and learn task-specific information, and this operation is finely performed at the layer level. Task-specific prompt tokens are learned by the corresponding task ground-truth back-propagation to update independently for each different task during training. The forward process of testing is the same as that of training. Task-specific prompt tokens for each task follow the same pipeline above, and interact with other shared parameters in the encoder separately.

\subsection{Efficient Multi-task Features Fusion}
\label{sec:3_3}
 Features from encoder are vital for some low-level tasks, such as boundary detection. To address the computational cost and quadratic complexity of the Transformer, a common strategy is to downsample the feature map to a lower resolution and then handle the multi-scale features from different transformer encoder layers. InvPT \cite{invpt} decoder designs a multi-scale features from encoder aggregation method, but these aggregate features are just multi-scale instead of multi-task and multi-scale features from encoder. In our task-specific prompts transformer encoder, task-specific prompts encoder corresponding task features. Multi-task features fusion mechanism is necessary, which can play the role of information complementation between different tasks.
 
 In order to better aggregate multi-task features, we apply aggregation weight matrix $W \in \mathbb{R}^{4 \times T}$ (fixed or learnable) for different $T$ tasks features $\mathrm{F}_i^t$ ($t=1,2, \ldots, T$) to obtain $i$-th scale  multi-task aggregation features $\mathrm{F}_i$ as follows: 
% \begin{equation}
% \mathrm{F}_i = \sum_{t=1}^T\left\mathrm{~W}_{t, i} \otimes \mathrm{F}_i^t\right
% \end{equation}

\begin{equation}
\mathbf{F}_i = \sum_{t=1}^T \left( \mathbf{W}_{t, i} \otimes \mathbf{F}_i^t \right)
\end{equation}

where $i$=1,2,3,4,and $\otimes$ means element-wise multiplication after broadcast operation. Our
experimental results show that aggregation weight matrix full of fixed-weight $\frac{\mathit{1}}{\mathit{T}}$ achieves a reasonably high level of performance. We carried out an enumeration experiment focusing on fixed weight cases and observed that the absence of encoder information for any task leads to performance drops. Furthermore, the complementarity of information between tasks is crucial, and more details can be found in \cref{sec:4_2}.

\subsection{Multi-head Transformer Decoder}
\label{sec:3_4}
Our MTL model uses multi-task features from encoder as input for Multi-head Transformer Decoder. In the encoder stage, we obtain enough task-specific information. The purpose of the decoder is to interact and refine task-specific features.

For the decoder part, we adopted the design structure of Inverted Pyramid Transformer Decoder \cite{invpt}, which is mainly composed of a multi-task up-transformer block that is applied to gradually increases the spatial resolution of the feature map and also perform multi-task features interaction to refine all tasks.

% The multi-task features after three stages up-transformer block that branches out into the task-specific heads. Finally, outputs pass through the 1 × 1 convolution layer of the task-specific channel number to get the final prediction.

\subsection{Loss Function}
After the multi-task decoder heads output the prediction results, we apply task-specific loss functions for each task to computer loss with ground truth. For pixel-by-pixel classification tasks: semantic segmentation, human parsing, saliency detection, and boundary detection, the cross-entropy loss is uniformly used for optimization. L1-Loss is used for depth estimation and surface normals estimation. The whole model can be end-to-end optimized. We set $\lambda_{\text{seg}}$=1.0, 
$\lambda_{\text{depth}}$=1.0, 
$\lambda_{\text{normals}}$= 10.0, 
$\lambda_{\text{edge}}$=50.0, 
$\lambda_{\text{part\_seg}}$=2.0, and 
$\lambda_{\text{sal}}$=5.0. $\lambda$ represents the weight of the loss corresponding to each task:
%\vspace{-0.1cm}
\begin{equation}
\begin{aligned}
\mathcal{L} & =\lambda_{\text {seg }} \mathcal{L}_{\text {seg }}+\lambda_{\text {depth }} \mathcal{L}_{\text {depth }}+\lambda_{\text {normals }} \mathcal{L}_{\text {normals }} \\
& +\lambda_{\text {edge }} \mathcal{L}_{\text {edge }}+\lambda_{\text {partseg }} \mathcal{L}_{\text {partseg }}+\lambda_{\text {sal }} \mathcal{L}_{\text {sal }}
\end{aligned}
\end{equation}

We use the exact same loss function weights as those in InvPT \cite{invpt}, without making any adjustments.

\section{Experiments}

\subsection{Experimental Setup}
\noindent\textbf{Datasets.}
Following previous works, we use NYUD-v2 and PASCAL-Context datasets for performance evaluation.
NYUD-v2 \cite{nyud} contains 795 training images and 654 testing images for indoor scenes, including four tasks: semantic segmentation, monocular depth estimation, surface normal estimation, and object boundary detection. PASCAL-Context \cite{pascal} contains 4998 training images and 5105 testing images. It labels for five tasks: semantic segmentation, human parts parsing, monocular depth estimation, surface normal estimation, and object boundary detection. We conduct experiments on both datasets for performance evaluation.\\
\noindent\textbf{Evaluation metric.}
Semantic segmentation and human parts parsing tasks both are segmentation tasks and are evaluated with the mean Intersection over Union (IoU) and monocular depth estimation task using the Root Mean Square Error (RMSE) as the evaluation metric. The surface normal estimation task uses mean angular error (mErr) as the evaluation metric, saliency detection is evaluated with the maximum F-measure (maxF), and the boundary detection task uses the optimal dataset-scale F-measure (odsF) score as the evaluation metric.

\noindent\textbf{Implementation details.} 
We mainly perform ablation studies using Vision Transformer(including Vit-L, Vit-B) pre-trained on ImageNet-21K \cite{imagenet} as the transformer encoder on the NYUD-v2 dataset. We set the training iterations of the model to 40k on the NYUD-v2 dataset and the PASCAL-Context dataset, both with a batch size of 6 using NVIDIA A40 GPU. For the optimization, Adam optimizer is used, and the learning rate is set to 2$\times10^{-5}$ with a weight decay rate of 1$\times10^{-6}$. A polynomial learning rate scheduler is used.

\begin{table}[h]
\begin{subtable}{0.5\textwidth}
\centering
\scalebox{0.9}{
	\begin{tabular}{@{}lllll@{}}
		\toprule
		Method        & \multicolumn{1}{c}{\begin{tabular}[c]{@{}c@{}}Semseg\\ (IoU)$\uparrow$\end{tabular}} & \multicolumn{1}{c}{\begin{tabular}[c]{@{}c@{}}Depth\\ (RMSE)$\downarrow$\end{tabular}} & \multicolumn{1}{c}{\begin{tabular}[c]{@{}c@{}}Normal\\ (mErr)$\downarrow$\end{tabular}} & \multicolumn{1}{c}{\begin{tabular}[c]{@{}c@{}}Boundary\\ (odsF)$\uparrow$\end{tabular}} \\ \midrule
		Cross-Stitch \cite{crossstitch} &  \multicolumn{1}{c}{36.34}                                                                          & \multicolumn{1}{c}{0.6290}                                                                           &  \multicolumn{1}{c}{20.88}                                                                            &  \multicolumn{1}{c}{76.38}                                                                             \\
		PAP \cite{pap}          &  \multicolumn{1}{c}{36.72}                                                                          & \multicolumn{1}{c}{0.6178}                                                                           & \multicolumn{1}{c}{20.82}                                                                            & \multicolumn{1}{c}{76.42}                                                                              \\
		PSD \cite{psd}         &  \multicolumn{1}{c}{36.69}                                                                          & \multicolumn{1}{c}{0.6246}                                                                           & \multicolumn{1}{c}{20.87}                                                                            & \multicolumn{1}{c}{76.42}                                                                               \\
		PAD-Net \cite{PAD-Net}         &  \multicolumn{1}{c}{36.61}                                                                          & \multicolumn{1}{c}{0.6270}                                                                           & \multicolumn{1}{c}{20.85}                                                                            & \multicolumn{1}{c}{76.38}                                                            \\
		MTI-Net \cite{mti-Net}       &  \multicolumn{1}{c}{45.97}                                                                          & \multicolumn{1}{c}{0.5365}                                                                           & \multicolumn{1}{c}{20.27}                                                                            & \multicolumn{1}{c}{77.86}                                                                               \\
		ATRC \cite{atrc}         &  \multicolumn{1}{c}{46.33}                                                                          & \multicolumn{1}{c}{0.5363}                                                                           & \multicolumn{1}{c}{20.18}                                                                            & \multicolumn{1}{c}{77.94}                                                                          \\
   		InvPT \cite{invpt}      &  \multicolumn{1}{c}{53.56}                                                                          & \multicolumn{1}{c}{0.5183}                                                                           & \multicolumn{1}{c}{19.04}                                                                            & \multicolumn{1}{c}{78.10}                                            \\
		TaskPrompter \cite{ye2023taskprompter}      &  \multicolumn{1}{c}{55.30}                                                                          & \multicolumn{1}{c}{0.5152}                                                                           & \multicolumn{1}{c}{18.47}                                                                            & \multicolumn{1}{c}{\textbf{78.20}}                                            \\
		\textbf{Ours}         & \multicolumn{1}{c}{\textbf{55.39}}                                                   & \multicolumn{1}{c}{\textbf{0.4961}}                                                 & \multicolumn{1}{c}{\textbf{18.44}}                                                    & \multicolumn{1}{c}{77.50}                                                     \\ \bottomrule
	\end{tabular}}
	\caption*{(a) State-of-the-art comparison on NYUD-v2.}
  \label{tab:sota nyud}
\end{subtable}
\begin{subtable}{0.5\textwidth}
    \centering
	\scalebox{0.78}{
		\begin{tabular}{@{}llllll@{}}
			\toprule
			Method           & \begin{tabular}[c]{@{}l@{}}Semseg\\ (IoU)$\uparrow$\end{tabular} & \begin{tabular}[c]{@{}l@{}}Parsing\\ (IoU)$\uparrow$\end{tabular} & \begin{tabular}[c]{@{}l@{}}Saliency\\ (maxF)$\uparrow$\end{tabular} & \begin{tabular}[c]{@{}l@{}}Normal\\ (mErr)$\downarrow$\end{tabular} & \begin{tabular}[c]{@{}l@{}}Boundary\\ (odsF)$\uparrow$\end{tabular} \\ \midrule
			ASTMT \cite{astmt}      & \multicolumn{1}{c}{68.00}       & \multicolumn{1}{c}{61.10}        & \multicolumn{1}{c}{65.70}            & \multicolumn{1}{c}{14.70}        & \multicolumn{1}{c}{72.40}         \\
			PAD-Net \cite{PAD-Net}    &    \multicolumn{1}{c}{53.60}       & \multicolumn{1}{c}{59.60}        & \multicolumn{1}{c}{65.80}            & \multicolumn{1}{c}{15.30}        & \multicolumn{1}{c}{72.50}       \\
			MTI-Net \cite{mti-Net}    &     \multicolumn{1}{c}{61.70}       & \multicolumn{1}{c}{60.18}        & \multicolumn{1}{c}{84.78}            & \multicolumn{1}{c}{14.23}        & \multicolumn{1}{c}{70.80}      \\
			ATRC \cite{atrc}       &     \multicolumn{1}{c}{62.69}       & \multicolumn{1}{c}{59.42}        & \multicolumn{1}{c}{84.70}            & \multicolumn{1}{c}{14.20}        & \multicolumn{1}{c}{72.96}     \\
			ATRC-ASPP \cite{atrc}  &     \multicolumn{1}{c}{63.60}       & \multicolumn{1}{c}{60.23}        & \multicolumn{1}{c}{83.91}            & \multicolumn{1}{c}{14.30}        & \multicolumn{1}{c}{70.86}        \\
			ATRC-BMTAS \cite{atrc} &     \multicolumn{1}{c}{67.67}       & \multicolumn{1}{c}{62.93}        & \multicolumn{1}{c}{82.29}            & \multicolumn{1}{c}{14.24}        & \multicolumn{1}{c}{72.42}          \\
			InvPT \cite{invpt}        &    \multicolumn{1}{c}{79.03}       & \multicolumn{1}{c}{67.61}        & \multicolumn{1}{c}{84.81}            & \multicolumn{1}{c}{14.15}        & \multicolumn{1}{c}{73.00}        \\
   TaskPrompter \cite{ye2023taskprompter}        &    \multicolumn{1}{c}{80.89}       & \multicolumn{1}{c}{68.89}        & \multicolumn{1}{c}{84.83}            & \multicolumn{1}{c}{13.72}        & \multicolumn{1}{c}{73.50}        \\
			\textbf{Ours}       & \multicolumn{1}{c}{\textbf{81.48}}        & \multicolumn{1}{c}{\textbf{70.64}}        & \multicolumn{1}{c}{\textbf{84.86}}         & \multicolumn{1}{c}{\textbf{13.69}}        & \multicolumn{1}{c}{\textbf{74.80}}          \\ \bottomrule
	\end{tabular}}
	\caption*{(b) State-of-the-art comparison on PASCAL-Context.}
  \label{tab:sota pascal}
\end{subtable}
   \captionsetup{belowskip=-8pt}
	\caption{State-of-the-art comparison on NYUD-v2 (a) and PASCAL-Context (b). Our method significantly outperforms the previous state-of-the-arts methods.}
 %\vspace{-5mm}
  \label{tab:sota}
\end{table}

%%\vspace{-5mm}
 \subsection{Comparisons With State-of-the-art Methods}
We compare our method with the state-of-the-art methods, including PAD-Net, MTI-Net, ATRC, InvPT and TaskPrompter. The results on the NYUD-v2 and Pascal-Context datasets are reported in Table \ref{tab:sota} (a) and Table \ref{tab:sota} (b), respectively. On NYUD-v2, the performance improvement for semantic segmentation, depth estimation, and surface normal estimation is significant, while for boundary detection, it is comparable with state-of-the-art methods. All tasks, except boundary detection, achieve state-of-the-art results on the NYUD-v2 dataset in our experimental results. As discussed in previous works \cite{sener2018multi,uncertainty}, one possible reason for the difference in performance gain is the task competition problem in training. The reason for task competition is also confirmed in our supplementary material. When each task does not share the encoder, the performance of boundary detection is almost the same as that of InvPT \cite{invpt}. Although the performance of boundary detection in Table \ref{tab:sota} has declined, overall, our results are still state-of-the-art. To enhance the overall performance, we present the performance in Table \ref{tab:sota}. On Pascal-Context, the improvements in semantic segmentation, human parsing, saliency detection, surface normal estimation, and boundary detection are all significant.
% These results validate the effectiveness of our proposed approach. 
% We further visualize the results of different methods in \cref{qualitative} on NYUD-v2. The results demonstrate that our method more closely approximates the ground truth.
% , further validating the effectiveness of task-specific prompts learning for holistic scene understanding. 

 \subsection{Ablation Studies}

 \begin{table}[]
 \begin{subtable}{0.5\textwidth}
 \centering
\scalebox{0.85}{
\begin{tabular}{@{}cccccc@{}}
\toprule
Task Prompt  & \begin{tabular}[c]{@{}c@{}}Semseg\\ (IoU)$\uparrow$\end{tabular} & \begin{tabular}[c]{@{}c@{}}Depth\\ (RMSE)$\downarrow$\end{tabular} & \begin{tabular}[c]{@{}c@{}}Normal\\ (mErr)$\downarrow$\end{tabular} & \begin{tabular}[c]{@{}c@{}}Boundary\\ (odsF)$\uparrow$\end{tabular} \\ \midrule
       % &                       & 53.56                                                       & 0.5183                                                     & 19.04                                                        & 78.10                                                         \\
                              & 53.56                                                       & 0.5183                                                     & 19.04                                                        & \textbf{78.10}                                                         \\

{\checkmark}                                & \textbf{55.39}                                                      & \textbf{0.4961}                                                      & \textbf{18.44}                                                      & {77.50}                                                         \\ \bottomrule
\end{tabular}}
	% \caption{Ablation for Task prompt \& Prompt cross-attention on NYUD-v2}
    \caption*{(a) Task prompt ablation on NYUD-v2.}
  \label{tab:ablation_prompt}
\end{subtable}

%\begin{table}[]
\begin{subtable}{0.5\textwidth}
\centering
\scalebox{0.75}{
\begin{tabular}{@{}cccccc@{}}
\toprule
Task Prompt  & \begin{tabular}[c]{@{}c@{}}Semseg\\ (IoU)$\uparrow$\end{tabular} & \begin{tabular}[c]{@{}c@{}}Parsing\\ (IoU)$\uparrow$\end{tabular} &
\begin{tabular}[c]{@{}c@{}}Saliency\\ (maxF)$\uparrow$\end{tabular}  &
\begin{tabular}[c]{@{}c@{}}Normal\\ (mErr)$\downarrow$\end{tabular} & 
\begin{tabular}[c]{@{}c@{}}Boundary\\ (odsF)$\uparrow$\end{tabular} 

\\ \midrule
       % &                       & 53.56                                                       & 0.5183                                                     & 19.04                                                        & 78.10                                                         \\
                              & 79.03                                                       & 67.61                                                     & 84.81                                                        & 14.15                         & 73.00                                 \\

{\checkmark}                                & \textbf{81.48}                                                      & \textbf{70.64}                                                      & \textbf{84.86}                                                       & \textbf{13.69}  &\textbf{74.80}                                                         \\ \bottomrule
\end{tabular}}
	% \caption{Ablation for Task prompt \& Prompt cross-attention on NYUD-v2}
    \caption*{(b) Task prompt ablation on PASCAL-Context.}
  \label{tab:ablation_prompt_pas}
\end{subtable}
\captionsetup{belowskip=-7pt}
    \caption{Task prompt ablation on on NYUD-v2 (a) and PASCAL-Context (b). Our  proposed task-specific prompt significantly enhances multi-task performance.}
  \label{tab:ablation_prompt}
   %\vspace{-11mm}
\end{table}

\noindent\textbf{The effectiveness of task-specific prompts.} 
To verify the effectiveness of our proposed task prompt learning method, we conduct ablation studies on both NYUD-v2 and PASCAL-Context. The introduction of task prompts significantly enhances multi-task performance, as demonstrated in Table \ref{tab:ablation_prompt} (a) and Table \ref{tab:ablation_prompt} (b). The performance improves because task-specific prompt tokens facilitate learning independent embedding spaces for each task.
 % by acquiring task-specific features at every layer within the encoder
Such a method decouples the characteristics of different tasks at the layer level, providing a finer and more flexible representation of various tasks. In the following section, we conduct comprehensive experiments to evaluate the effectiveness of task-specific prompts and derive numerous valuable insights, such as the positions of task-specific prompts hold greater significance than the number of prompt tokens per task in each layer and introducing shared prompt tokens results in unfavorable performance, among other conclusions.
% the initialization of prompts has a more significant impact than anticipated; and introducing shared prompt tokens results in unfavorable performance, among other conclusions.

\begin{figure}[h]
	\includegraphics[width=0.47\textwidth]{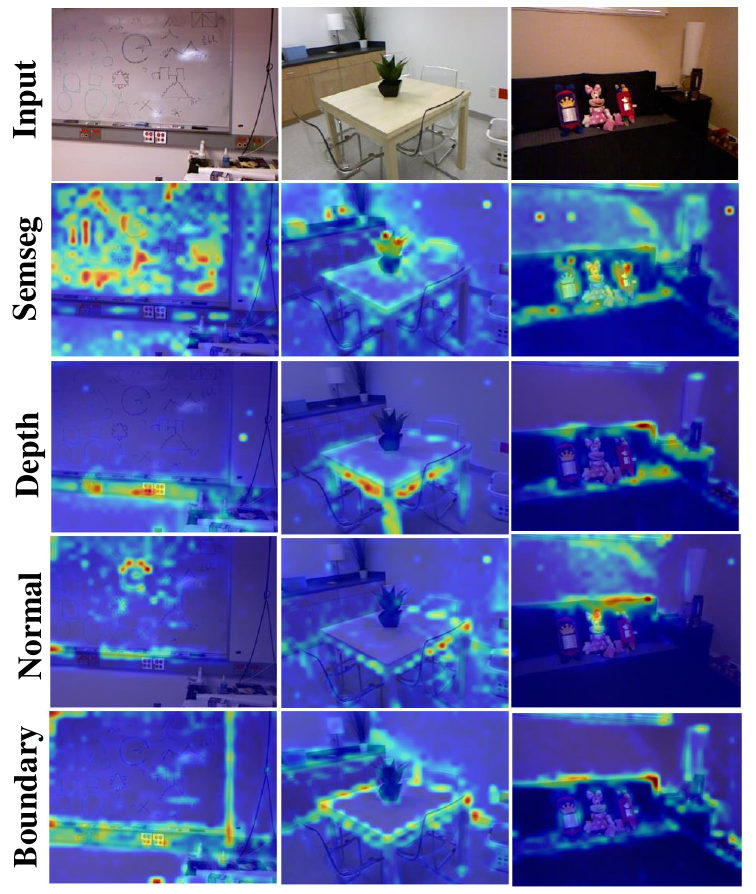}
	% \caption{Visualization of task-specific prompts attention map. It can be observed that task-specific prompts can attend to distinct spatial positions of the patch tokens, which indicates that the task prompts can effectively learn task-specific representations from the interaction with the image patch tokens.}
         \captionsetup{belowskip=-10.5pt}
 \caption{Visualizing the attention maps for task-specific prompts demonstrates their ability to focus on distinct spatial positions within patch tokens. This observation indicates that task prompts effectively learn task-specific representations through their interactions with image patch tokens.}
	%\label{fig:galaxy}%what use?
\vspace{-10pt}
	\label{attn}
\end{figure}

\noindent\textbf{The effectiveness of efficient multi-task features fusion.}
\label{sec:4_2}
As shown in Table \ref{tab:ablation_encoder_feature}, we explore a variety of efficient multi-task encoder features fusion strategies.
% , including fixed weight, learnable weight, cross-task attention, etc. 
In theory, employing learnable weights
% initialized with fixed values 
and utilizing cross-task attention offers a broader latent space than a fixed-weight strategy, potentially leading to better performance through an optimal solution. However, our experimental results indicate that more complex fusion strategies tend to yield poorer performance, suggesting that the role of this block is both critical and sensitive. Complex fusion strategies can increase instability in the model optimization process for our multi-objective optimization problem \cite{sener2018multi}, leading to suboptimal performance. Moreover, multi-task features from encoder are essential, particularly for low-level tasks such as boundary detection. We conducted an enumeration experiment concentrating on fixed-weight scenarios and observed that the lack of encoder information for any task results in performance decreasing. Additionally, the complementarity of information between tasks is vital.

\begin{table}[]
\begin{subtable}{0.5\textwidth}
\scalebox{0.8}{\begin{tabular}{@{}cccccc@{}}
\toprule
\multicolumn{1}{l}{Method}  & \begin{tabular}[c]{@{}c@{}}Semseg\\ (IoU)$\uparrow$\end{tabular} & \begin{tabular}[c]{@{}c@{}}Depth\\ (RMSE)$\downarrow$\end{tabular} & \begin{tabular}[c]{@{}c@{}}Normal\\ (mErr)$\downarrow$\end{tabular} & \begin{tabular}[c]{@{}c@{}}Boundary\\ (odsF)$\uparrow$\end{tabular} \\\midrule
\multicolumn{1}{l}{w/o fusion}                  & \multicolumn{1}{c}{53.47} & \multicolumn{1}{c}{0.5113} & \multicolumn{1}{c}{19.59} & \multicolumn{1}{c}{69.50} \\
\multicolumn{1}{l}{fixed weight}         & \multicolumn{1}{c}{\textbf{55.39}} & \multicolumn{1}{c}{\textbf{0.4961}} & \multicolumn{1}{c}{\textbf{18.44}} & \multicolumn{1}{c}{\textbf{77.50}} \\
\multicolumn{1}{l}{learnable weight}     & \multicolumn{1}{c}{55.18} & \multicolumn{1}{c}{0.4998} & \multicolumn{1}{c}{18.59} & \multicolumn{1}{c}{\textbf{77.50}} \\
\multicolumn{1}{l}{cross-task attention} & \multicolumn{1}{c}{55.16} & \multicolumn{1}{c}{0.5066} & \multicolumn{1}{c}{18.60} & \multicolumn{1}{c}{77.30} \\ \hline
\end{tabular}}
    \caption*{(a) Ablation for efficient multi-task features fusion strategy on NYUD-v2.}
  \label{tab:ablation_encoder_feature}
\end{subtable}
% \hfill
\begin{subtable}{0.5\textwidth}
%\begin{table}[]
\scalebox{0.8}{\begin{tabular}{@{}cccccc@{}}
\toprule
\multicolumn{1}{l}{Fixed weight}  & \begin{tabular}[c]{@{}c@{}}Semseg\\ (IoU)$\uparrow$\end{tabular} & \begin{tabular}[c]{@{}c@{}}Depth\\ (RMSE)$\downarrow$\end{tabular} & \begin{tabular}[c]{@{}c@{}}Normal\\ (mErr)$\downarrow$\end{tabular} & \begin{tabular}[c]{@{}c@{}}Boundary\\ (odsF)$\uparrow$\end{tabular} \\\midrule
\multicolumn{1}{l}{$0.25,0.25,0.25,0.25$ }          & \multicolumn{1}{c}{\textbf{55.39}} & \multicolumn{1}{c}{\textbf{0.4961}} & \multicolumn{1}{c}{\textbf{18.44}} & \multicolumn{1}{c}{\textbf{77.50}} \\ 
\multicolumn{1}{l}{$0.00,0.25,0.25,0.25$}         & \multicolumn{1}{c}{54.59} & \multicolumn{1}{c}{0.4993} & \multicolumn{1}{c}{18.72} & \multicolumn{1}{c}{77.00} \\
\multicolumn{1}{l}{$0.25,0.00,0.25,0.25$}     & \multicolumn{1}{c}{55.18} & \multicolumn{1}{c}{0.4980} & \multicolumn{1}{c}{18.70} & \multicolumn{1}{c}{77.00} \\
\multicolumn{1}{l}{$0.25,0.25,0.00,0.25$} & \multicolumn{1}{c}{55.02} & \multicolumn{1}{c}{0.5019} & \multicolumn{1}{c}{18.65} & \multicolumn{1}{c}{77.40} \\ 
\multicolumn{1}{l}{$0.25,0.25,0.25,0.00$}     & \multicolumn{1}{c}{55.14} & \multicolumn{1}{c}{0.5031} & \multicolumn{1}{c}{18.68} & \multicolumn{1}{c}{77.20} \\
\multicolumn{1}{l}{$0.25,0.25,0.00,0.00$}     & \multicolumn{1}{c}{54.86} & \multicolumn{1}{c}{0.5000} & \multicolumn{1}{c}{18.63} & \multicolumn{1}{c}{77.10} \\
\multicolumn{1}{l}{$0.25,0.00,0.00,0.00$}     & \multicolumn{1}{c}{54.59} & \multicolumn{1}{c}{0.5012} & \multicolumn{1}{c}{18.74} & \multicolumn{1}{c}{77.30} \\
\multicolumn{1}{l}{$0.00,0.00,0.00,0.00$}     & \multicolumn{1}{c}{53.47} & \multicolumn{1}{c}{0.5113} & \multicolumn{1}{c}{19.59} & \multicolumn{1}{c}{69.50} \\
\hline
\end{tabular}}
    \caption*{(b) Ablation for efficient multi-task features fusion fixed weight on NYUD-v2. For the NYUD-v2 dataset consisting of four tasks, using a fixed weight of $\frac{1}{T}=0.25$ and arranged in the sequence of semantic segmentation, depth estimation, surface normal estimation, and boundary detection.} 
  \label{tab:fusion_weight}
\end{subtable}
%\captionsetup{belowskip=-2mm}
%\captionsetup{-420mm}
 \caption{Ablation for efficient multi-task features fusion module. We explore a variety of efficient multi-task encoder features fusion strategies (a) and achieve the best performance with fixed values. Concentrating on fixed-weight scenarios (b) and observed that the lack of encoder information for any task results in performance decreasing.}
 \vspace{-8mm}
 \label{tab:ablation_encoder_feature}
\end{table}
%\vspace{-3mm}
% \begin{figure*}[h] %图片放置的还不够对称
% 	\includegraphics%[width=20cm]
% 	[scale=0.45]{Figure/pas383.pdf}
% 	\caption{Qualitative comparison on Pascal-Context}
% 	%\label{fig:galaxy}%what use?
% 	\label{quapas}
% \end{figure*}
\noindent\textbf{Task-specific prompts vs. task-unified prompts.} 
In our approach, we propose that task-specific prompts encode the induced prior knowledge for each task, promoting task-specific representation learning. 
% Meanwhile, the transformer with shared parameters facilitates connections between different tasks.
Besides using task-specific prompts, we also design experiments by introducing task-unified prompts, which are the same for all tasks and do not update independently. The task-specific prompts and task-unified prompts can be used in the following ways: 1) task-unified prompts only; 2) task-unified prompts and task-specific prompts are concatenated together; 3) task-unified prompts and task-specific prompts are blended with an element-wise summation; 4) task-specific prompt only. 5) task-unified prompts generated by fusing task-specific prompts with cross-prompt attention operation; task-specific prompts and fused task-unified prompts are blended with an element-wise summation.
% 5) task-specific prompts and task-unified prompts generated by task-specific prompts after cross-prompt attention operation are blended with an element-wise summation. 
It is worth noting that 3) will also lead to task-specific prompts, but the optimization in 3) is different from that in 4) because the task-unified prompts are supervised by four different tasks while the task-specific prompts are supervised by the corresponding task only. Results in Table \ref{tab:ablation_unify} show that task-unified prompts constantly undermine the overall performance, compared with our method with task-specific prompts only, where $n$ and $m$ mean the number of task-unified prompts and task-specific prompts. There are two possible reasons for the poor performance of task-unified prompts: i) it may be possible to exist the task-unified prompts for fewer tasks, but difficult to exist the task-unified prompts for all four tasks; ii) the encoder aims at explicitly decoupling the features of different tasks.
% , and the decoder will fuse features of different tasks.
Introducing the task-unified prompts in the encoder goes against the feature decoupling for different tasks. %To verify the first assumption, we only learn common prompts XXX and XXX tasks, and our initial results show that indeed the common prompts help the performance improvement for fewer tasks. XXXXXXXXXXXXXXXXXXXXXXXToBeVerifiedByExperimentsXXXXXXXXXXXXXX.%

\begin{table}[]
\scalebox{0.7}{
\centering
\begin{tabular}{@{}ccccccc@{}}
\toprule
$m$    & $n$                  &method                 &  \begin{tabular}[c]{@{}c@{}}Semseg\\ (IoU)$\uparrow$\end{tabular} & \begin{tabular}[c]{@{}c@{}}Depth\\ (RMSE)$\downarrow$\end{tabular} & \begin{tabular}[c]{@{}c@{}}Normal\\ (mErr)$\downarrow$\end{tabular} & \begin{tabular}[c]{@{}c@{}}Boundary\\ (odsF)$\uparrow$\end{tabular}  \\ \midrule
5  & 0  & / &  \textbf{55.39} & 0.4961 & \textbf{18.44} & 77.50  \\
5  & 1 & concat  & 55.23 & \textbf{0.4935} & 18.65 & 76.90\\5  & 5 & concat & 54.45 & 0.4940 & 18.63 & 77.40 \\5  & 5 & add  & 54.82 & 0.5001 & 18.58 & 77.40 \\
0  & 5 &  /  & 53.45 & 0.5042 & 18.81 & 77.30 \\
0  & 20 &  / & 53.43 & 0.5055 & 18.72 & 77.40 \\
 \hline
5  & 5 & cross-prompt attention & 55.38 & 0.5016 & 18.56 & \textbf{77.80} \\ \bottomrule
\end{tabular}}
%\captionsetup{belowskip=-10.5pt}
    \caption{Performance by different combinations of the task-unified and task-specific prompts on NYUD-v2. The task-specific prompts and task-unified prompts can be blended in the above ways where $n$ and $m$ mean the number of task-unified prompts and task-specific prompts. Results show that task-unified prompts always undermine the overall performance, compared with task-specific prompts only.}
    \vspace{-15pt}
  \label{tab:ablation_unify}
  %%\vspace{-40pt}
\end{table}

\noindent\textbf{The positions of task-specific prompts.}
In our implementation, the transformer encoder contains 24 layers in total.
% , and task-specific prompts can also be inserted into a single layer or multiple layers. 
We denote the input as layer 1. In Table \ref{tab:ablation for depth}, we report the results with prompts inserted at multiple layers where $i-j$ means the task-specific prompts are used from the $i$-th layer to the $j$-th layer. First, we can see that inserting task-specific prompts into more layers always boosts the performance for semantic segmentation and depth estimation, but for surface normal estimation and boundary detection, more layers with visual prompts do not necessarily bring about the performance gain. Considering the trade-off between performance and computational costs, we use prompts for 12 layers. Further, we propose inserting prompts at different positions, and it shows that the performance is better by inserting the visual prompts at deeper layers of the encoder on NYUD-v2.  %It is also reasonable because the low-level features of different tasks should be more related compared with high-level features.  
Thus we choose to insert the visual prompts for layers between 13-24 in all our ablation experiments on NYUD-v2. 
\begin{table}[h]
	\scalebox{0.8}{
		\begin{tabular}{@{}clllllll@{}}
			\toprule
			Layers with prompts & \multicolumn{1}{c}{\begin{tabular}[c]{@{}c@{}}Semseg\\ (IoU)$\uparrow$\end{tabular}} & \multicolumn{1}{c}{\begin{tabular}[c]{@{}c@{}}Depth\\ (RMSE)$\downarrow$\end{tabular}} & \multicolumn{1}{c}{\begin{tabular}[c]{@{}c@{}}Normal\\ (mErr)$\downarrow$\end{tabular}} & \multicolumn{1}{c}{\begin{tabular}[c]{@{}c@{}}Boundary\\ (odsF)$\uparrow$\end{tabular}} \\ \midrule
                                 			w/o prompt                 &  \multicolumn{1}{c}{53.56}                           & \multicolumn{1}{c}{0.5183}       & \multicolumn{1}{c}{19.04}        & \multicolumn{1}{c}{\textbf{78.10}}         \\
			1-6                  & \multicolumn{1}{c}{54.24}                      & \multicolumn{1}{c}{0.4989} & \multicolumn{1}{c}{18.72} & \multicolumn{1}{c}{77.40}    \\
			1-12                 &  \multicolumn{1}{c}{54.96}                           & \multicolumn{1}{c}{0.4948}       & \multicolumn{1}{c}{18.72}        & \multicolumn{1}{c}{77.40}         \\
   			1-18                 &  \multicolumn{1}{c}{55.45}                           & \multicolumn{1}{c}{0.4929}       & \multicolumn{1}{c}{18.51}        & \multicolumn{1}{c}{77.60}         \\
               			1-24                 &  \multicolumn{1}{c}{\textbf{55.80}}                           & \multicolumn{1}{c}{\textbf{0.4898}}       & \multicolumn{1}{c}{18.63}        & \multicolumn{1}{c}{77.60}         \\
\midrule
			1-12                 &  \multicolumn{1}{c}{54.96}                           & \multicolumn{1}{c}{0.4948}       & \multicolumn{1}{c}{18.72}        & \multicolumn{1}{c}{77.40}         \\
			13-24                 & \multicolumn{1}{c}{55.39}                           & \multicolumn{1}{c}{0.4961}       & \multicolumn{1}{c}{\textbf{18.44}}        & \multicolumn{1}{c}{77.50}         \\   
   \bottomrule
	\end{tabular}}
	\caption{The performance with different prompts inserting positions on NYUD-v2. Inserting task-specific prompts into more layers always boosts the performance for semantic segmentation and depth estimation, but for surface normal estimation and boundary detection, more layers with prompts do not necessarily bring about the performance gain.}
 %\vspace{-23pt}
 \label{tab:ablation for depth}
\end{table}

\vspace{-12pt}

\noindent\textbf{The number of task-specific prompts per layer.} As shown in Table \ref{tab:ablation for prompt num}, the performance is not always improved when the number of prompt tokens increases. Performance decreases as the number of prompt tokens increases to a certain threshold, and the computational demand rises significantly. Moreover, each task appears to favors a distinct number of task-specific prompts. Depth, normal and boundary accuracy achieved the best results when token number $N$=5. Unlike these tasks, semantic segmentation achieves the best performance at $N$=50. However, the change in performance is relatively minor. For all experiments with task-specific prompts, even with only one prompt for each task, the performance is always better than without task-specific prompts.
% , further validating the effectiveness of task-specific prompts. 
\begin{table}[h]
	\scalebox{0.8}{
		\begin{tabular}{@{}clllllll@{}}
			\toprule
			Prompt Number & \multicolumn{1}{c}{\begin{tabular}[c]{@{}c@{}}Semseg\\ (IoU)$\uparrow$\end{tabular}} & \multicolumn{1}{c}{\begin{tabular}[c]{@{}c@{}}Depth\\ (RMSE)$\downarrow$\end{tabular}} & \multicolumn{1}{c}{\begin{tabular}[c]{@{}c@{}}Normal\\ (mErr)$\downarrow$\end{tabular}} & \multicolumn{1}{c}{\begin{tabular}[c]{@{}c@{}}Boundary\\ (odsF)$\uparrow$\end{tabular}} \\ \midrule
                                    			w/o prompt                &  \multicolumn{1}{c}{53.56}                           & \multicolumn{1}{c}{0.5183}       & \multicolumn{1}{c}{19.04}        & \multicolumn{1}{c}{\textbf{78.10}}         \\
			1            & \multicolumn{1}{c}{55.27}                           & \multicolumn{1}{c}{0.4990}      & \multicolumn{1}{c}{18.81}       & \multicolumn{1}{c}{77.00}         \\
			5           & \multicolumn{1}{c}{55.39}                           & \multicolumn{1}{c}{\textbf{0.4961}}      & \multicolumn{1}{c}{\textbf{18.44}}       & \multicolumn{1}{c}{77.50}         \\
			10           & \multicolumn{1}{c}{55.19}                           & \multicolumn{1}{c}{\textbf{0.4961}}      & \multicolumn{1}{c}{18.58}       & \multicolumn{1}{c}{77.60}         \\
			50           & \multicolumn{1}{c}{\textbf{55.61}}                           & \multicolumn{1}{c}{0.4972}      & \multicolumn{1}{c}{18.66}       & \multicolumn{1}{c}{77.10}         \\ 
   			100           & \multicolumn{1}{c}{55.18}                           & \multicolumn{1}{c}{0.5001}      & \multicolumn{1}{c}{18.61}       & \multicolumn{1}{c}{77.40}         \\
      			200           & \multicolumn{1}{c}{55.38}                           & \multicolumn{1}{c}{0.4971}      & \multicolumn{1}{c}{18.69}       & \multicolumn{1}{c}{77.30}         \\ 
   \bottomrule
	\end{tabular}}
	\caption{Ablation for Prompt Number on NYUD-v2. As the number of prompt tokens increases, the performance does not always get better.}
 \vspace{-20pt}
 \label{tab:ablation for prompt num}
\end{table}

\noindent\textbf{Influences of different backbones.}
To validate the generalization of our task-specific prompts, we also use different backbones, and the results are shown in Table \ref{tab:ablation for backbone}. We can see that with different backbones, our task-specific prompts always boost the performance for various scene-understanding tasks. So our task-specific prompts strategy can be compatible with other backbones. 
\begin{table}[h]
\scalebox{0.85}{
	\begin{tabular}{@{}ccccccc@{}}
		\toprule
		\multicolumn{1}{l}{Backbone} & \multicolumn{1}{c}{\begin{tabular}[c]{@{}c@{}}Semseg\\ (IoU)$\uparrow$\end{tabular}} & \multicolumn{2}{c}{\begin{tabular}[c]{@{}c@{}}Depth\\ (RMSE)$\downarrow$\end{tabular}} & \multicolumn{1}{c}{\begin{tabular}[c]{@{}c@{}}Normal\\ (mErr)$\downarrow$\end{tabular}} & \multicolumn{2}{c}{\begin{tabular}[c]{@{}c@{}}Boundary\\ (odsF)$\uparrow$\end{tabular}} \\ \midrule
		%running
		\multicolumn{1}{l}{InvPT(Vit-B)}                        & \multicolumn{1}{c}{50.30}                                                                           & \multicolumn{2}{c}{0.5367}                                                       & 19.00                                                                           & \multicolumn{2}{c}{77.60}                                                          \\ 
  		%running
		\multicolumn{1}{l}{Ours(Vit-B)}                        & 51.22                                                                           & \multicolumn{2}{c}{0.5301}                                                       & 18.78                                                                           & \multicolumn{2}{c}{76.90}                                                          \\ 
  		\multicolumn{1}{l}{InvPT(Vit-L)}                        & 53.56                                                                           & \multicolumn{2}{c}{0.5183}                                                       & 19.04                                                                            & \multicolumn{2}{c}{\textbf{78.10}}                                                          \\ 
		\multicolumn{1}{l}{Ours(Vit-L)}                        & \textbf{55.39}                                                                           & \multicolumn{2}{c}{\textbf{0.4961}}                                                       & \textbf{18.44}                                                                            & \multicolumn{2}{c}{77.50}                                                          \\  
  \bottomrule
	\end{tabular}}
	\caption{Ablation for Backbone on NYUD-v2. With different backbones, our task-specific prompts always boost the performance for different scene-understanding tasks.}
  \vspace{-10pt}
 \label{tab:ablation for backbone}
\end{table}

\noindent\textbf{The task specialty of the different task-specific prompts.} 
To show the specialty of the different task-specific prompts, we replace the original corresponding task prompt tokens with the same trained task prompt tokens in inference. 
%We explored in our default setting $D_{12,24}$ and half/all task prompt means replace original corresponding task prompt tokens by  $D_{12,24}$ or $D_{18,24}$.
Results in Table \ref{tab:ablation for task-specific} show that the performance drops by replacing the task-specific prompts with prompt tokens learned for the other tasks. The reason is that different task-specific prompts are pretty different from each other. It is worth noting that when we replace all prompts with task-specific prompts, such as those for semantic segmentation, the corresponding performance for the given task also changes due to the cross-task feature fusion module. To show whether the task prompts learn task-specific representation on patch tokens more intuitively, we visualize the attention map values between task prompts and patch tokens in the Encoder module as shown in Fig. \ref{attn}. The attention map values are highly related to each task's particularity, demonstrating that the task prompts can effectively encode task-specific representations. 
\begin{table}[]
\scalebox{0.8}{
\begin{tabular}{@{}lllll@{}}
\toprule
\multicolumn{1}{l}{Model} &  \multicolumn{1}{c}{\begin{tabular}[c]{@{}c@{}}Semseg\\ (IoU)$\uparrow$\end{tabular}} & \multicolumn{1}{c}{\begin{tabular}[c]{@{}c@{}}Depth\\ (RMSE)$\downarrow$\end{tabular}} & \multicolumn{1}{c}{\begin{tabular}[c]{@{}c@{}}Normal\\ (mErr)$\downarrow$\end{tabular}} & \multicolumn{1}{c}{\begin{tabular}[c]{@{}c@{}}Boundary\\ (odsF)$\uparrow$\end{tabular}}  \\ \midrule
    \multicolumn{1}{l}{task-specific prompt}   & \multicolumn{1}{c}{\textbf{55.39}} & \multicolumn{1}{c}{0.4961} & \multicolumn{1}{c}{18.44} & \multicolumn{1}{c}{77.50} \\ 
 \multicolumn{1}{l}{segmentation prompt}                  & \multicolumn{1}{c}{55.35}                           & \multicolumn{1}{c}{0.5596}      & \multicolumn{1}{c}{20.58}       & \multicolumn{1}{c}{77.50}       \\
%\multicolumn{1}{c}{half semseg prompt}            & \multicolumn{1}{c}{55.38}                           & \multicolumn{1}{c}{0.4947}      & \multicolumn{1}{c}{18.58}       & \multicolumn{1}{c}{77.70}   \\
%\midrule
\multicolumn{1}{l}{depth prompt}            & \multicolumn{1}{c}{48.30}                           & \multicolumn{1}{c}{\textbf{0.4950}}      & \multicolumn{1}{c}{18.76}       & \multicolumn{1}{c}{77.00}       \\
%\multicolumn{1}{c}{half depth prompt}            & \multicolumn{1}{c}{54.90}                           & \multicolumn{1}{c}{0.4937}      & \multicolumn{1}{c}{18.48}       & \multicolumn{1}{c}{77.60} \\ \midrule
\multicolumn{1}{l}{normal prompt}            & \multicolumn{1}{c}{40.85}                           & \multicolumn{1}{c}{0.5533}      & \multicolumn{1}{c}{\textbf{18.43}}       & \multicolumn{1}{c}{76.60}       \\
%\multicolumn{1}{c}{half normal prompt}            & \multicolumn{1}{c}{54.74}                           & \multicolumn{1}{c}{0.4996}      & \multicolumn{1}{c}{18.52}       & \multicolumn{1}{c}{77.60}\\
 %\midrule
\multicolumn{1}{l}{edge prompt}            & \multicolumn{1}{c}{52.88}                           & \multicolumn{1}{c}{0.5497}      & \multicolumn{1}{c}{20.15}       & \multicolumn{1}{c}{\textbf{77.60}}       \\
%\multicolumn{1}{c}{half edge prompt}            & \multicolumn{1}{c}{54.67}                           & \multicolumn{1}{c}{0.4988}      & \multicolumn{1}{c}{18.56}       & \multicolumn{1}{c}{77.60} \\
 \bottomrule
\end{tabular}}
\caption{Ablation for prompt task-specialty. The performance drops by replacing the task-specific prompts with prompt tokens learned for other tasks.}
%\vspace{-25pt}
 \label{tab:ablation for task-specific}
\end{table}
\begin{figure}[h]
	\includegraphics%[width=20cm]
	[scale=0.45]{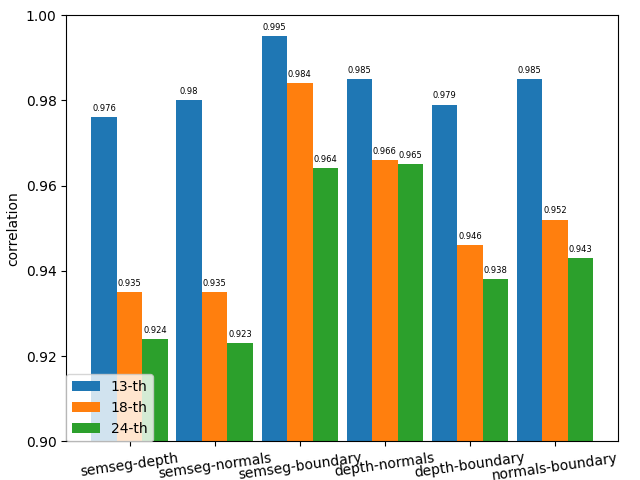}
  \captionsetup{belowskip=-11pt}
	\caption{Investigating the correlation of features corresponding to different tasks in $\{13,18,24\}-th$ layers output space. Such as semseg-depth to indicate the feature correlation between segmentation and depth estimation.}
	%\label{fig:galaxy}%what use?
         \vspace{-3mm}
	\label{fig:correlation}
\end{figure}

\noindent\textbf{The correlation of features corresponding to different tasks.} 
In our experiments, we introduce the task-specific prompts from the 13th to the 24th layer. We also calculate the average correlation for features corresponding to different tasks. We denote the out space features corresponding to the $k$-th layer as $F_k^A$ for task-A and denote the features of the $k$-th of the same image as $F_k^B$, then we calculate the cosine distance between $F_k^A$ and $F_k^B$ and do the average over all the test images. The results are shown in \cref{fig:correlation}. We can see that as the layer goes deeper, as we introduce more and more task-specific prompts, the features corresponding to different tasks become more and more uncorrelated.
\vspace{-3mm}
% \noindent\textbf{Visualization of task-specific features corresponding to different tasks.} 
% Our task-specific prompts help us learn task-specific features. Thus we use XXX strategy to visualize which image contents contribute more to different tasks. The results are shown in XXXX. We can see that XXXXXXXXXXXXXX.

\section{Conclusion}
\vspace{-2mm}
This paper presents a task-specific prompts-boosted transformer for holistic scene understanding, where the task-specific prompts are fed into different input layers of the transformer encoder for different tasks while the parameters of the transformer layer are shared across different tasks. Then we fuse task-specific features of different tasks with a task-fusion module and feed the fused feature into the transformer decoder scene prediction. Extensive experiments show that our method achieves state-of-the-art performance, which validates the effectiveness of our approach.

\noindent\textbf{Acknowledgement.} 
The work was supported by NSFC \#61932020, \#62172279, Program of Shanghai Academic Research Leader, and “Shuguang Program" supported by Shanghai Education Development Foundation and Shanghai Municipal Education Commission.

%%%%%%%%% REFERENCES
{\small
\bibliographystyle{ieee_fullname}
\bibliography{egbib}
}

\end{document}

\title{TSP-Transformer: Task-Specific Prompts Boosted Transformer for Holistic Scene Understanding}

\author{Shuo Wang$^{1}$\quad
Jing Li$^{2}$ \quad
Zibo Zhao$^{1}$ \quad
Dongze Lian$^{3}$ \\
Binbin Huang$^{1}$ \quad
Xiaomei Wang$^{4}$ \quad
Zhengxin Li$^{1}$ \quad
Shenghua Gao$^{1,5,6}$\footnote[2]{} \quad
\\
$^{1}$ShanghaiTech University \quad
$^{2}$Xiaohongshu Inc.\quad
$^{3}$National University of Singapore \\
$^{4}$Fudan University\quad
$^{5}$Shanghai Engineering Research Center of Intelligent Vision and Imaging\\
$^{6}$Shanghai Engineering Research Center of Energy Efficient and Custom AI IC\\
{\tt\small   \{wansghuo2022, zhaozb, liandz, huangbb, lizhx, gaoshh\}@shanghaitech.edu.cn\\ \tt\small lijing1@alumni.shanghaitech.edu.cn, 17110240025@fudan.edu.cn}
}
% \and
\maketitle
\renewcommand{\tabcolsep}{1.5ex}

\section*{Appendices}
\subsection*{Appendix. A \quad More Implementation Details}
\noindent\textbf{Data Processing.} For a fair comparison with ATRC \cite{atrc}, InvPT \cite{invpt} and TaskPrompter \cite{ye2023taskprompter}, we follow their data processing pipeline. On PASCAL-Context \cite{pascalcontext}, we pad the image to the size of 512 × 512, while on NYUD-v2 \cite{nyud}, we randomly crop the input image to the size of 448 × 576. We use typical data augmentation including random scaling, cropping, horizontal flipping and color jittering.

% \subsection*{Appendix. B \quad Implementation Details of Encoder Feature Fusion}
\noindent\textbf{Implementation Details of Encoder Feature Fusion.}
For ViT \cite{vit} encoders, we follow InvPT \cite{invpt} implementation choosing 3 layers based on the depth and unfolding their output spatially, and then use transposed convolution to upsample the resolution of feature maps to match the spatial resolution in the corresponding decoder stage before the further transformation. Specifically, for ViT-base encoder, using the output token sequences of layer 3, 6, and 9, while for ViT-large encoder using output token sequences of layer 6, 12, and 18. The encoder feature fusion module procures multi-task encoder features of different scales from the preceding layers. The kernel size and stride of the transposed convolution for the feature at the first scale are 4, and those at the second scale are 2. The fused coarse multi-task encoder feature and the fine multi-task feature after intermediate supervision participate in cross-attention operations to forward in the decoder.

\begin{table}[h]
\scalebox{0.93}{
\begin{tabular}{@{}clllllll@{}}
\toprule
\multicolumn{1}{l}{Method} & \multicolumn{1}{c}{\begin{tabular}[c]{@{}c@{}}Semseg\\ (IoU)$\uparrow$\end{tabular}} & \multicolumn{1}{c}{\begin{tabular}[c]{@{}c@{}}Depth\\ (RMSE)$\downarrow$\end{tabular}} & \multicolumn{1}{c}{\begin{tabular}[c]{@{}c@{}}Normal\\ (mErr)$\downarrow$\end{tabular}} & \multicolumn{1}{c}{\begin{tabular}[c]{@{}c@{}}Boundary\\ (odsF)$\uparrow$\end{tabular}} \\ \midrule
\multicolumn{1}{l}{unshared}            & \multicolumn{1}{c}{54.03}                           & \multicolumn{1}{c}{0.5121}      & \multicolumn{1}{c}{18.86}       & \multicolumn{1}{c}{\textbf{78.00}}         \\
\multicolumn{1}{l}{shared}             & \multicolumn{1}{c}{\textbf{55.39}}                           & \multicolumn{1}{c}{\textbf{0.4961}}      & \multicolumn{1}{c}{\textbf{18.44}}       & \multicolumn{1}{c}{77.50}         \\ \bottomrule
\end{tabular}}
\caption{Ablation for shared encoder on NYUD-v2. Performance with the shared encoder is better for all the tasks except for boundary detection.}
 \label{tab:ablation for shared}
 %\vspace{-23pt}
\end{table}

\subsection*{Appendix. B \quad More Experimental Results and Analysis}
\noindent\textbf{Shared encoder vs. unshared encoder.} 
We advocate that different tasks are closely related, and a shared encoder makes various tasks share the same low-level features (layers 1-12) and different but task-specific high-level features(layers 13-24). Other than using a shared encoder, we also report the performance based on task-specific encoders where different encoders with task-specific prompts are learned for different tasks. It means that our encoder has a different branch for each task, and the different tasks in the encoder stage are completely independent, without any interaction between tasks and no shared parameters in the encoder. 

Results on NYUD-v2 with different encoder design strategies are reported in Table \ref{tab:ablation for shared}. It shows that the results with the shared encoder are better for all the tasks except for boundary detection. The possible reason is that semantic segmentation, depth estimation, and surface normal estimation are more closely related tasks. In contrast, its relation to other tasks is not that strong for boundary detection. The difference in performance gain is the task competition problem in training. Thus learned task-specific encoder may not be a wrong choice for boundary detection. 

\begin{table}[h]
\scalebox{0.9}{
\begin{tabular}{@{}lllll@{}}
\toprule
\multicolumn{1}{l}{method} & \multicolumn{1}{c}{\begin{tabular}[c]{@{}c@{}}SemSeg\\ (IoU)$\uparrow$\end{tabular}} & \multicolumn{1}{c}{\begin{tabular}[c]{@{}c@{}}Depth\\ (RMSE)$\downarrow$\end{tabular}} & \multicolumn{1}{c}{\begin{tabular}[c]{@{}c@{}}Normal\\ (mErr)$\downarrow$\end{tabular}} & \multicolumn{1}{c}{\begin{tabular}[c]{@{}c@{}}Boundary\\ (odsF)$\uparrow$\end{tabular}}  \\ \midrule
\multicolumn{1}{l}{zeros}  &  \multicolumn{1}{c}{\textbf{55.39}}&  \multicolumn{1}{c}{\textbf{0.4961}}& \multicolumn{1}{c}{\textbf{18.44}} & \multicolumn{1}{c}{\textbf{77.50}} \\
\multicolumn{1}{l}{random} & \multicolumn{1}{c}{54.31} & \multicolumn{1}{c}{0.5069} & \multicolumn{1}{c}{18.67} & \multicolumn{1}{c}{77.40} \\
\multicolumn{1}{l}{ones}   & \multicolumn{1}{c}{54.61} & \multicolumn{1}{c}{0.4962} & \multicolumn{1}{c}{18.66} & \multicolumn{1}{c}{\textbf{77.50}} \\ \bottomrule
\end{tabular}}
	\caption{Performance with different prompt initialization strategies on NYUD-v2. It’s impressive that our default initialization zeros, generally works the best.}
  \label{tab:initialization}
\end{table}

\noindent\textbf{Prompts Initialization.} 
Prompt tuning first emerged in the field of NLP, and the research on prompt initialization is an important field. Visual prompt tuning \cite{vpt} also studies the initialization method of the prompts, but it is only a tuning setting. Compared with Visual prompt tuning, the model parameters that can be optimized vary greatly in our multi-task learning method. The conclusions drawn above do not necessarily apply to our multi-task transformer network.

We compare the performance using the above initialization strategy against the default zeros initialization in Table \ref{tab:initialization}.  As shown in Table \ref{tab:initialization}, it’s impressive that our default initialization zeros, works the best in general. 

\begin{figure}[h]
	\includegraphics%[width=20cm]
	[scale=0.55]{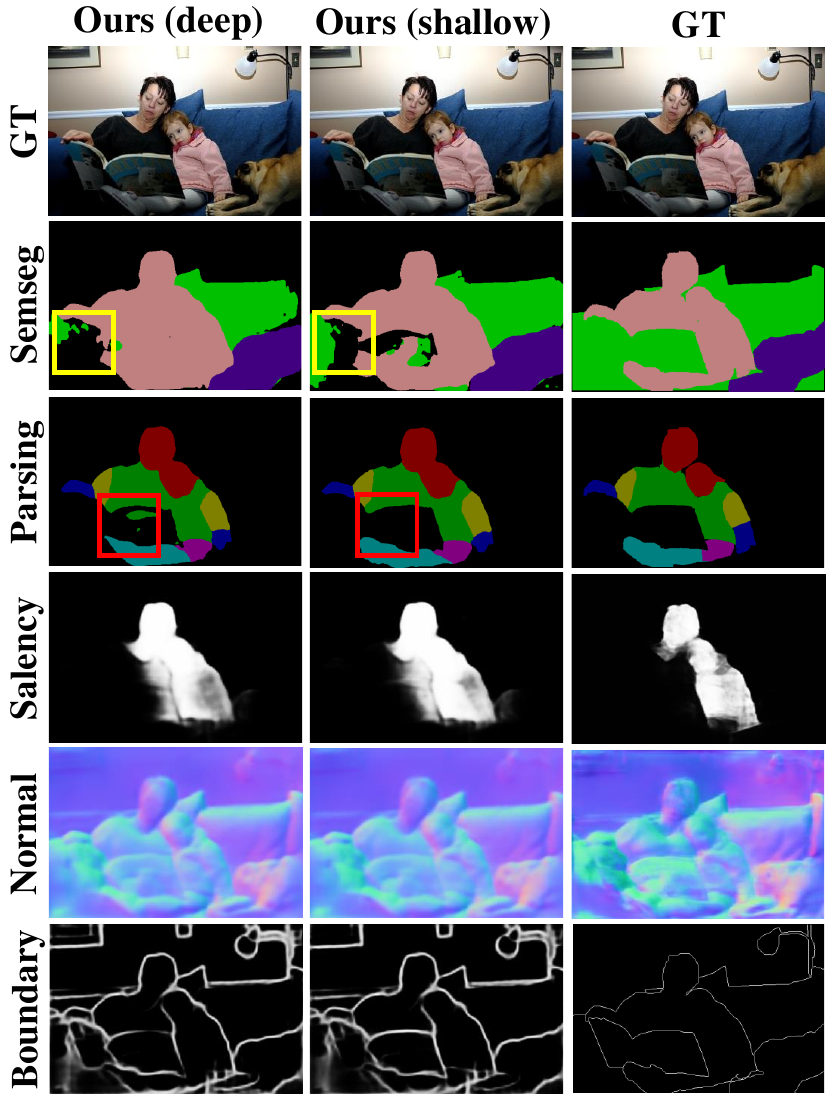}
   %\captionsetup{belowskip=-13pt}
	\caption{Qualitative analysis of different model variants (shallow: $1-12$ and deep: $13-24$) on PASCAL-Context. Results of different model variants are shown by Table \ref{tab:ablation for dataset}.}
	%\label{fig:galaxy}%what use?
 %\vspace{-12pt}
	\label{pas270}
\end{figure}
\begin{table}[t]
\begin{subtable}{0.5\textwidth}
	\scalebox{0.8}{
		\begin{tabular}{@{}clllllll@{}}
			\toprule
			Layers with prompts & \multicolumn{1}{c}{\begin{tabular}[c]{@{}c@{}}Semseg\\ (IoU)$\uparrow$\end{tabular}} & \multicolumn{1}{c}{\begin{tabular}[c]{@{}c@{}}Depth\\ (RMSE)$\downarrow$\end{tabular}} & \multicolumn{1}{c}{\begin{tabular}[c]{@{}c@{}}Normal\\ (mErr)$\downarrow$\end{tabular}} & \multicolumn{1}{c}{\begin{tabular}[c]{@{}c@{}}Boundary\\ (odsF)$\uparrow$\end{tabular}} \\ \midrule
                                 			w/o prompt                 &  \multicolumn{1}{c}{53.56}                           & \multicolumn{1}{c}{0.5183}       & \multicolumn{1}{c}{19.04}        & \multicolumn{1}{c}{\textbf{78.10}}         \\
			1-12                 &  \multicolumn{1}{c}{54.96}                           & \multicolumn{1}{c}{\textbf{0.4948}}       & \multicolumn{1}{c}{18.72}        & \multicolumn{1}{c}{77.40}         \\

               			13-24                 &  \multicolumn{1}{c}{\textbf{55.39}}                           & \multicolumn{1}{c}{0.4961}       & \multicolumn{1}{c}{\textbf{\textbf{18.44}}}        & \multicolumn{1}{c}{77.50}          \\  
               			% 1-24                 &  \multicolumn{1}{c}{\textbf{55.80}}                           & \multicolumn{1}{c}{\textbf{0.4898}}       & \multicolumn{1}{c}{18.63}        & \multicolumn{1}{c}{77.60}          \\  
   \bottomrule
	\end{tabular}}
	\caption*{(a) The performance with prompts positions on NYUD-v2}
  %\vspace{-2mm}
 %\label{tab:ablation for dataset1}
\end{subtable}
 \hfill
\begin{subtable}{0.5\textwidth}
	\scalebox{0.7}{
		\begin{tabular}{@{}cllllllll@{}}
			\toprule
			Layers with prompts  & \begin{tabular}[c]{@{}l@{}}Semseg\\ (IoU)$\uparrow$\end{tabular} & \begin{tabular}[c]{@{}l@{}}Parsing\\ (IoU)$\uparrow$\end{tabular} & \begin{tabular}[c]{@{}l@{}}Saliency\\ (maxF)$\uparrow$\end{tabular} & \begin{tabular}[c]{@{}l@{}}Normal\\ (mErr)$\downarrow$\end{tabular} & \begin{tabular}[c]{@{}l@{}}Boundary\\ (odsF)$\uparrow$\end{tabular} \\  \midrule
                                 			w/o prompt                 &  \multicolumn{1}{c}{79.03}       & \multicolumn{1}{c}{67.61}        & \multicolumn{1}{c}{84.81}            & \multicolumn{1}{c}{14.15}        & \multicolumn{1}{c}{73.00}        \\
			1-12                 &  \multicolumn{1}{c}{\textbf{81.48}}                           & \multicolumn{1}{c}{\textbf{70.64}}       & \multicolumn{1}{c}{\textbf{84.86}}        & \multicolumn{1}{c}{\textbf{13.69}}    & \multicolumn{1}{c}{\textbf{74.80}}      \\

               			13-24                 & \multicolumn{1}{c}{80.64}        & \multicolumn{1}{c}{69.53}        & \multicolumn{1}{c}{84.67}         & \multicolumn{1}{c}{13.83}        & \multicolumn{1}{c}{74.20}              \\
               			% 1-24                 & \multicolumn{1}{c}{\textbf{80.64}}        & \multicolumn{1}{c}{\textbf{69.53}}        & \multicolumn{1}{c}{\textbf{84.67}}         & \multicolumn{1}{c}{\textbf{13.83}}        & \multicolumn{1}{c}{\textbf{74.60}}              \\
      
   \bottomrule
	\end{tabular}}
	\caption*{(b) The performance with prompts positions on PASCAL-Context}
  % \label{tab:ablation for dataset1}
 \end{subtable}
 \captionsetup{belowskip=-10pt}
 \caption{The performance with prompts positions on two different datasets. Unlike the NYUD-v2 dataset, the placement depth of the task prompts substantially impacts performance, particularly for higher-level scene understanding tasks such as Semseg and Parsing on PASCAL-Context.}
 %\vspace{-8mm}
 \label{tab:ablation for dataset}
\end{table}

\begin{table*}[h!]
\begin{subtable}{1.0\textwidth}
	\scalebox{0.9}{
		\begin{tabular}{@{}cclllllllll@{}}
			\toprule
			Layers with prompts & Prompt token numbers & \multicolumn{1}{c}{\begin{tabular}[c]{@{}c@{}}Semseg\\ (IoU)$\uparrow$\end{tabular}} & \multicolumn{1}{c}{\begin{tabular}[c]{@{}c@{}}Depth\\ (RMSE)$\downarrow$\end{tabular}} & \multicolumn{1}{c}{\begin{tabular}[c]{@{}c@{}}Normal\\ (mErr)$\downarrow$\end{tabular}} & \multicolumn{1}{c}{\begin{tabular}[c]{@{}c@{}}Boundary\\ (odsF)$\uparrow$\end{tabular}}  & \multicolumn{1}{c}{\begin{tabular}[c]{@{}c@{}}GFlops\\
            (G)\end{tabular}}    & \multicolumn{1}{c}{\begin{tabular}[c]{@{}c@{}}Number of parameters\\
            (M)\end{tabular}}
            \\ \midrule
                                 			w/o prompt (InvPT)       & 0         &  \multicolumn{1}{c}{53.56}                           & \multicolumn{1}{c}{0.5183}       & \multicolumn{1}{c}{19.04}        & \multicolumn{1}{c}{\textbf{78.10}}                & \multicolumn{1}{c}{597.67}
                                    & \multicolumn{1}{c}{402.09}      \\
			24        & 1          & \multicolumn{1}{c}{53.97}                      & \multicolumn{1}{c}{0.5038} & \multicolumn{1}{c}{18.63} & \multicolumn{1}{c}{77.50}                                      & \multicolumn{1}{c}{654.17}
                                    & \multicolumn{1}{c}{402.10}  \\
   			13-24        & 5         &  \multicolumn{1}{c}{55.39}                           & \multicolumn{1}{c}{0.4961}       & \multicolumn{1}{c}{\textbf{18.44}}        & \multicolumn{1}{c}{77.50}                                        & \multicolumn{1}{c}{1146.24}
                                    & \multicolumn{1}{c}{402.34}     \\
               			1-12    & 5             & \multicolumn{1}{c}{54.96}                           & \multicolumn{1}{c}{0.4948}       & \multicolumn{1}{c}{18.72}        & \multicolumn{1}{c}{77.40}                                      & \multicolumn{1}{c}{1681.11}
                                    & \multicolumn{1}{c}{402.34}          \\

			1-24      & 5           &   \multicolumn{1}{c}{\textbf{55.80}}                           & \multicolumn{1}{c}{\textbf{0.4898}}       & \multicolumn{1}{c}{18.63}        & \multicolumn{1}{c}{77.60}                                          & \multicolumn{1}{c}{1685.13}
                                    & \multicolumn{1}{c}{402.58}     \\
   \bottomrule
	\end{tabular}}
	\caption*{The performance and compute cost with different prompts inserting positions and prompt token numbers for each task on NYUD-v2.}
 %\vspace{-23pt}
 %\label{tab:ablation for depth}
\end{subtable}

\begin{subtable}{1.0\textwidth}
	\scalebox{0.85}{
		\begin{tabular}{@{}cclllllllll@{}}
			\toprule
			Layers with prompts & Prompt token numbers & \begin{tabular}[c]{@{}l@{}}Semseg\\ (IoU)$\uparrow$\end{tabular} & \begin{tabular}[c]{@{}l@{}}Parsing\\ (IoU)$\uparrow$\end{tabular} & \begin{tabular}[c]{@{}l@{}}Saliency\\ (maxF)$\uparrow$\end{tabular} & \begin{tabular}[c]{@{}l@{}}Normal\\ (mErr)$\downarrow$\end{tabular} & \begin{tabular}[c]{@{}l@{}}Boundary\\ (odsF)$\uparrow$\end{tabular}   & \multicolumn{1}{c}{\begin{tabular}[c]{@{}c@{}}GFlops\\
            (G)\end{tabular}}    & \multicolumn{1}{c}{\begin{tabular}[c]{@{}c@{}}Number of parameters\\
            (M)\end{tabular}}\\ \midrule
                                 			w/o prompt (InvPT)        & 0         &  \multicolumn{1}{c}{79.03}       & \multicolumn{1}{c}{67.61}        & \multicolumn{1}{c}{84.81}            & \multicolumn{1}{c}{14.15}        & \multicolumn{1}{c}{73.00} 
                                    & \multicolumn{1}{c}{668.29}
                                    & \multicolumn{1}{c}{422.93}\\
			24        & 1          & \multicolumn{1}{c}{80.08}                      & \multicolumn{1}{c}{69.12} & \multicolumn{1}{c}{84.46} & \multicolumn{1}{c}{13.85}   & \multicolumn{1}{c}{74.10}                                     & \multicolumn{1}{c}{744.94}
                                    & \multicolumn{1}{c}{422.94}\\
			
   			13-24        & 5         &   \multicolumn{1}{c}{80.64}        & \multicolumn{1}{c}{69.53}        & \multicolumn{1}{c}{84.67}         & \multicolumn{1}{c}{13.83}        & \multicolumn{1}{c}{74.20}                                      & \multicolumn{1}{c}{1412.52}
                                    & \multicolumn{1}{c}{423.24} \\
               			1-12    & 5             &  \multicolumn{1}{c}{81.48}                           & \multicolumn{1}{c}{70.64}       & \multicolumn{1}{c}{84.86}        & \multicolumn{1}{c}{\textbf{13.69}}    & \multicolumn{1}{c}{\textbf{74.80}}                                      & \multicolumn{1}{c}{2138.61}
                                    & \multicolumn{1}{c}{423.24}\\

			1-24      & 5           &    \multicolumn{1}{c}{\textbf{81.63}}                           & \multicolumn{1}{c}{\textbf{70.69}}       & \multicolumn{1}{c}{\textbf{84.90}}        & \multicolumn{1}{c}{13.81}    & \multicolumn{1}{c}{74.70}                                          & \multicolumn{1}{c}{2143.65}
                                    & \multicolumn{1}{c}{423.54}   \\
   \bottomrule
	\end{tabular}}
	\caption*{The performance and compute cost with different prompts inserting positions and prompt token numbers for each task on PASCAL-Context.}
 %\vspace{-23pt}
 %\label{tab:ablation for depth}
\end{subtable}
 \caption{ On our most efficient model variant, which introduces only one task prompt token for each task on the last transformer encoder layer, the performance improvement is also significant, accompanied by a slight increase in the number of parameters and GFlops. It can also be seen that the shared vanilla layer placed in the shallow layer (layers with prompts: 24, 13-24) can drastically reduce the computational load in terms of GFlops, and there is almost a negligible increase in the number of parameters across all of our model variants.}
 %\vspace{-8mm}
 \label{tab:cost}
\end{table*}

\noindent\textbf{Dataset size and task numbers.}
We conduct extensive experiments on NYUD-v2 and PASCAL-
Context for performance evaluation, but our experiments reveal inconsistent patterns between the two datasets. %NYUD-v2 contains 795 training images and 654 testing images for indoor scenes including four tasks, while PASCAL-Context which provides labels for five tasks contains 4998 training images and 5105 test-ing images. 
Compared to the NYUD-v2 dataset, the PASCAL-Context dataset has a larger data size, a greater number of tasks, and a distinct data distribution. We believe these factors account for the discrepancies observed in some experimental results.

For NYUD-v2, there is no significant difference between the task prompts positions with the same layer range length in the shallow ($1-12$) and deep ($13-24$) layers of the encoder as shown in Table \ref{tab:ablation for dataset} (a). In fact, the results are slightly better in the task prompts positions with deeper layers. However, this is not the case with another PASCAL-Context dataset as shown in Table \ref{tab:ablation for dataset} (b). The positions has a more substantial impact on performance, particularly for certain higher-level scene understanding tasks such as Semseg and Parsing. A potential reason for this discrepancy is that high-level tasks demand more task-specific information, particularly as the size of data increases, task prompts positions with shallow layers leading to a finer decoupling of features from encoder.
\begin{figure}[t]
	\includegraphics%[width=30cm]
	[scale=0.3]{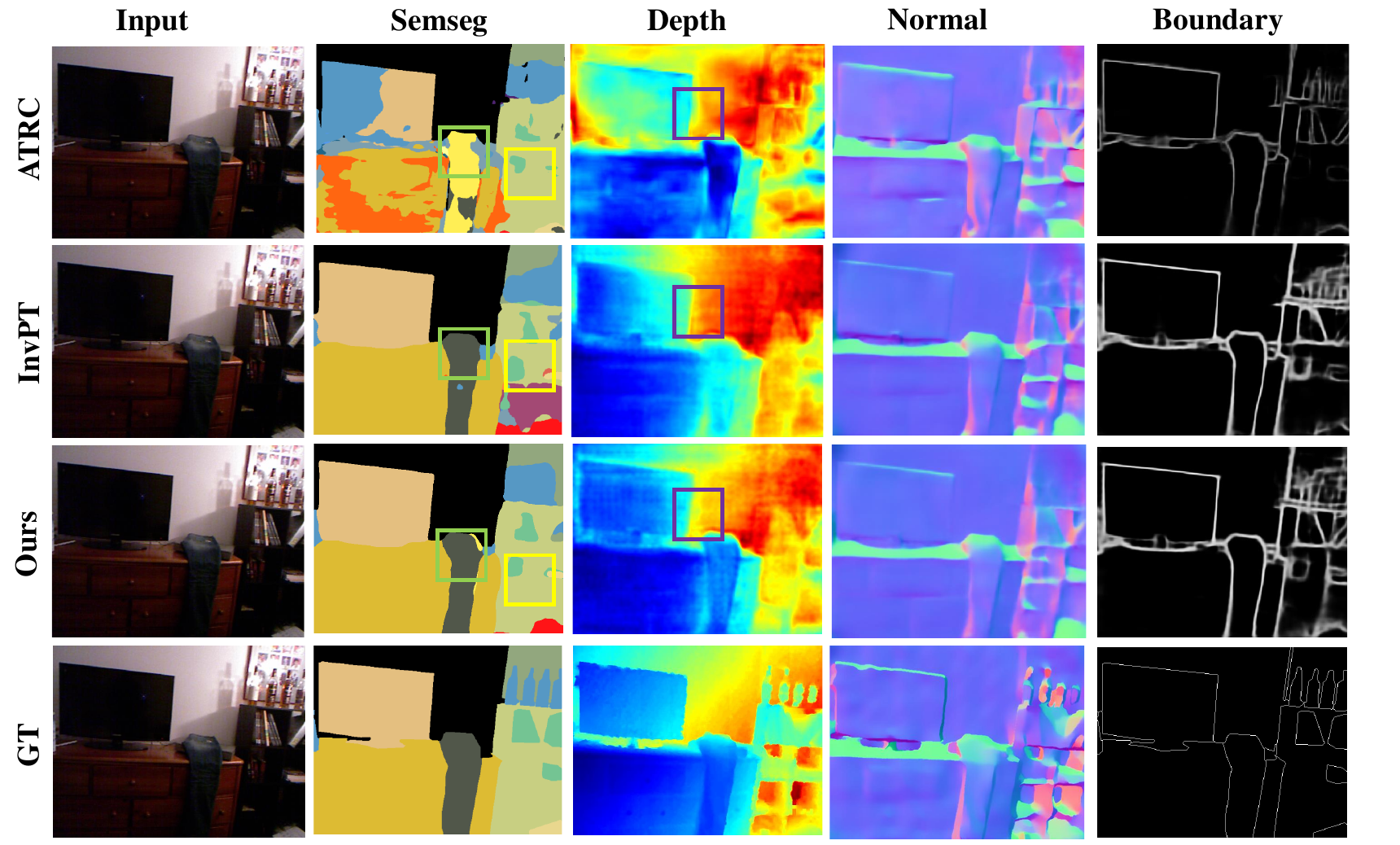}
	\caption{Qualitative comparison with state-of-the-art method on NYUD-v2. Our method generates more accurate predictions.}
	%\label{fig:galaxy}%what use?
	\label{qualitative}
\end{figure}
\begin{figure*}[h!]
	\includegraphics[width=\linewidth]{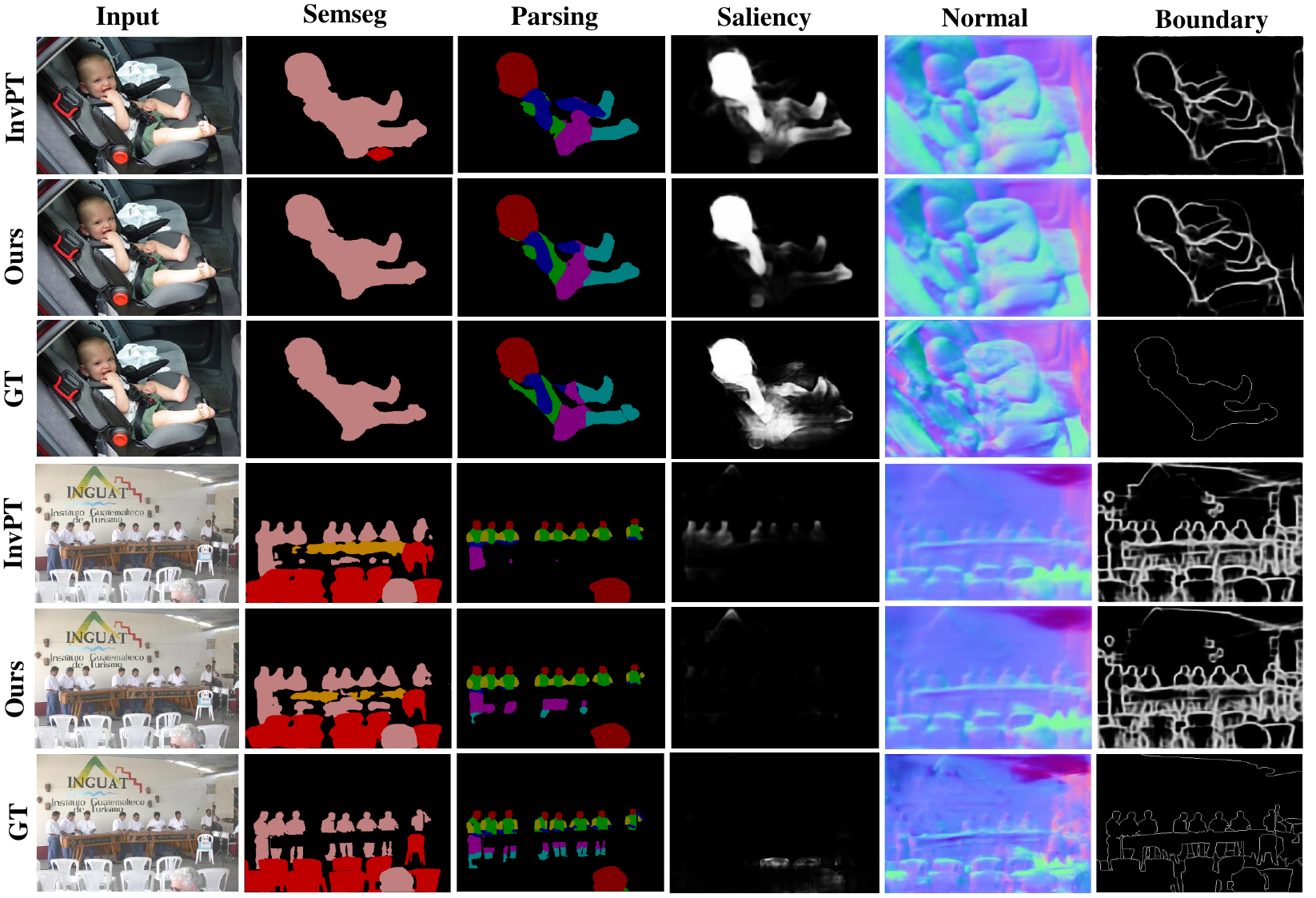}
   %\captionsetup{belowskip=-13pt}
	\caption{Qualitative comparison with state-of-the-art method on PASCAL-Context. Our method generates more accurate predictions.}
	%\label{fig:galaxy}%what use?
 %\vspace{-12pt}
	\label{pasmore}
\end{figure*}

\noindent\textbf{Computation cost.} Table \ref{tab:cost} shows the computation cost of our proposed model, detailing the GFLOPs and the number of model parameters across different model variants, along with corresponding performances on the NYUD-v2 and PASCAL-Context datasets. Our model variants demanding higher computational resources do not necessarily guarantee superior overall performance, indicating the existence of a trade-off between performance and computation. %image size 448 × 576 and 512 × 512 using a NVIDIA A40 GPU,using fvoce toolkit. 
% It is evident that an increase in the number of tasks correlates with an increase in the model’s parameters and computational requirements. This trend is primarily due to the increased number of the encoder and task prompt tokens interactions, the number of heads in the decoder stage, the number of tokens, and the larger image resolution required by the PASCAL-Context dataset. 
The most efficient model variant, which introduces only one task prompt token to each task on the last transformer encoder layer, demonstrates significant performance improvement with a modest increase in parameter count and GFLOPs. Furthermore, our results show that placing the shared vanilla layer in the shallow layer can drastically reduce the computational load in terms of GFLOPs. Moreover, across all our model variants, we observe only a nearly negligible increase in the number of parameters.

\noindent\textbf{More qualitative results.} We show more prediction results by our method and InvPT on the NYUD-v2 and PASCAL-Context dataset in Fig. \ref{qualitative} and Fig. \ref{pasmore}. It is clear that our method produces significantly better results than InvPT, especially on semantic segmentation, depth estimation and human parsing.




{\small
\bibliographystyle{ieee_fullname}
\bibliography{egbib}
}